\documentclass[10pt]{article}

\usepackage[T1]{fontenc}
\usepackage{arxiv}
\usepackage{amsmath}
\usepackage{amssymb}
\usepackage{algorithm}
\usepackage{algpseudocode}
\usepackage{booktabs}
\usepackage[section]{placeins}
\usepackage{framed}
\usepackage{multirow}
\usepackage{array}
\usepackage{graphicx}
\usepackage{xcolor}
\usepackage{tikz}
\usepackage{hyperref}
\usepackage{url}
\usetikzlibrary{arrows.meta,positioning,calc}

\newcommand{\algorithmframe}[2]{\vbox{\hrule height .8pt\kern 2pt
  #1\hbox{#2}\kern 2pt\hrule}}
\newenvironment{pagedalgorithm}[1]{%
  \def\LastFrameCommand{\algorithmframe{\hbox{\textbf{Algorithm \thealgorithm} (continued)}\kern 2pt\hrule\kern 2pt}}%
  
  \setlength{\OuterFrameSep}{7pt plus 1pt minus 1pt}%
  \MakeFramed{\FrameRestore}%
  \refstepcounter{algorithm}%
  \addcontentsline{loa}{algorithm}{\protect\numberline{\thealgorithm}#1}%
  \noindent\textbf{Algorithm \thealgorithm} #1\par\nobreak
  \hrule\kern 2pt\nobreak
}{\endMakeFramed}

\hypersetup{
  pdftitle={Low-Rank Weight Sharing for Neural Architecture Search},
  pdfauthor={Dhanush M and Ajin George Joseph},
  colorlinks=true,
  linkcolor=blue!60!black,
  citecolor=blue!60!black,
  urlcolor=blue!60!black
}

\title{Low-Rank Weight Sharing\\for Neural Architecture Search}
\renewcommand{\shorttitle}{Low-Rank Weight Sharing for Architecture Search}
\author{%
  \begin{tabular}[t]{@{}c@{\qquad}c@{}}
    \textbf{Dhanush M}\thanks{Corresponding author: Dhanush M.} &
    \textbf{Ajin George Joseph}\\
    \href{mailto:cs22b053@iittp.ac.in}{\texttt{cs22b053@iittp.ac.in}} &
    \href{mailto:ajin@iittp.ac.in}{\texttt{ajin@iittp.ac.in}}
  \end{tabular}\\[0.6em]
  Computer Science and Engineering\\
  Indian Institute of Technology Tirupati%
}
\date{}

\begin{document}
\raggedbottom
\maketitle

\begin{abstract}
Weight sharing makes neural architecture search less expensive by training candidate architectures within one network, where they reuse the parameters of components they have in common. However, a filter that performs well on features from one incoming edge may be less effective on features from another. We use low-rank weight sharing: candidates share the base weights, and selected convolutions learn a low-rank correction for each input source. We evaluate this design in two architecture search settings that use policy optimization. For image classification, the search policy proposes moves in the continuous representation learned by Neural Architecture Optimization (NAO), using the change in predicted accuracy as its reward. In the classification experiment, the search takes 1.48 A100 GPU-days, and the resulting model has 86.94M non-auxiliary parameters. When the official test split is evaluated after each epoch, accuracy peaks at 98.00\% and ends at 97.78\%. For semantic segmentation, the policy instead selects a discrete cell and receives a reward measured on labeled images. Under matched 500,000-update training, changing one edge raises best single-scale validation mean intersection over union (mIoU) from 73.942\% for our reproduced Auto-DeepLab-S to 74.285\%. With multi-scale and flip inference, the corresponding scores are 75.99\% and 76.32\%, although the selected cell performs worse on its lowest-scoring fifth of images.
\end{abstract}

\section{Introduction}

Designing convolutional architectures by training every candidate from scratch is expensive. In cell-based neural architecture search, each candidate defines a directed acyclic graph of operations and input connections that is repeated across the depth of the network. Even with a modest number of intermediate nodes and candidate operations, the number of valid encodings grows combinatorially into search spaces far too large to evaluate exhaustively. At the same time, fine-grained structural decisions matter: even a change in the source of a single convolutional input yields a distinct architecture that can affect accuracy, and verifying whether such an edit helps requires training both configurations under the same schedule. Repeating full stand-alone training runs across a large search space can demand hundreds to thousands of GPU-days~\cite{luo2018nao,real2019regularized}, making direct evaluation impractical under realistic compute budgets.

Within a cell, a connection determines which intermediate representation reaches a node, while the operation on that connection determines how the representation is transformed. Connectivity and operation choice can therefore be explored jointly, using feedback from earlier candidates to guide the search. Neural Architecture Optimization (NAO), for example, represents these choices in a continuous space and proposes changes that a learned predictor expects to improve performance~\cite{luo2018nao}. A promising move in that space, however, must still decode to a useful discrete architecture.

Weight sharing addresses the cost of training these candidates. Rather than maintaining a separate model for each design, a supernet trains sampled paths through a common pool of parameters, so an update to a shared component also affects other candidates that use it~\cite{pham2018enas}. This reuse can introduce interference. Consider two incoming connections whose features differ in scale or encode different visual patterns: an update that makes a shared operation better suited to one source may make it less suited to the other. Consequently, a candidate's score reflects both its architecture and the weights it inherits, and rankings under weight sharing need not persist after independent training~\cite{luo2019balanced}.

These issues arise in both classification and segmentation, although the meaning of better performance differs across the two tasks. On classification benchmarks, the usual goal is high overall accuracy. In semantic segmentation, by contrast, mean intersection over union (mIoU) gives equal weight to each class after pooling pixel counts across images. Consequently, class-averaged mIoU can improve even when the most difficult images are segmented no more accurately. A search criterion that is effective for classification therefore need not transfer directly to segmentation.

In this paper, we consider low-rank weight sharing as a way to retain parameter reuse while allowing an operation to specialize for different inputs. Candidates share the weights of common components; in selected convolutions, the effective weight is the sum of a shared base and a learned low-rank correction specific to the input source~\cite{hu2022lora}. We examine this principle under two search procedures based on proximal policy optimization (PPO)~\cite{schulman2017ppo}. For classification, PPO proposes moves in the continuous representation learned by NAO~\cite{luo2018nao}, whereas for segmentation, PPO selects discrete cells along a fixed Auto-DeepLab resolution path~\cite{liu2019autodeeplab}. Independent training changes the ordering of the shortlisted cells in both experiments. In the segmentation comparison, the selected cell improves class-averaged mIoU over our reproduced Auto-DeepLab-S under matched training conditions, yet its mean score on the lowest-scoring fifth of images declines.

\section{Related Literature}

Cell-based search methods differ in how they propose the next architecture. NASNet samples cells with a learned controller, whereas regularized evolution mutates cells selected from a population~\cite{zoph2018nasnet,real2019regularized}. Progressive NAS uses predicted accuracy to decide which partial cells to expand~\cite{liu2018pnas}. NAO instead moves the continuous encoding of a complete cell toward higher predicted accuracy and decodes the result~\cite{luo2018nao}. DARTS makes a different quantity continuous: it learns a mixture of operations on each edge before selecting a discrete cell~\cite{liu2019darts}.

How a candidate is proposed is distinct from how it is scored. ENAS and NAO-WS evaluate candidates with inherited weights~\cite{pham2018enas,luo2018nao}, and Bender et al.\ showed that a single shared model can identify promising candidates even without a learned controller~\cite{bender2018oneshot}. Nevertheless, Yu et al.\ found that the ordering under shared weights can differ substantially from the ordering after independent training~\cite{yu2020evaluating}. Two responses are to balance how often candidate paths are trained~\cite{luo2019balanced} or to divide the search space among several supernets so that fewer candidates share each set of weights~\cite{zhao2021fewshot}.

For semantic segmentation, the scope of the search also matters. Chen et al.\ searched for cells that combine features at several spatial scales~\cite{chen2018dense}, while Nekrasov et al.\ searched for compact decoders using auxiliary cells that provide supervision during training and are removed for inference~\cite{nekrasov2019auxiliary}. Auto-DeepLab extends the choice beyond the cell itself by jointly searching its operations and the sequence of spatial resolutions through the encoder~\cite{liu2019autodeeplab}.

In this paper, sharing is adjusted within a convolution: different input sources reuse a common filter but each receives its own low-rank correction~\cite{hu2022lora}. Both searches use proximal policy optimization~\cite{schulman2017ppo}, drawing on the idea behind conditional value at risk (CVaR) that an average can conceal poor outcomes in the lower tail~\cite{rockafellar2000cvar,tamar2015cvar}. What counts as a poor outcome differs by task. For classification, the policy proposes moves in NAO's continuous representation and gives priority to low-scoring moves in its update, without estimating an exact CVaR gradient. For segmentation, the policy selects a discrete cell on a fixed resolution path, and its reward accounts for the lowest-scoring images evaluated by that cell.

\section{Classification}
\label{sec:classification}

Discriminative image classification often depends on contextual relationships among visual features: a texture shared across categories may be more informative when interpreted alongside a larger shape. Connectivity governs which representations meet, whereas operations determine how they are transformed and combined. Because an operation's efficacy depends on its inputs, we search over both choices jointly. Through low-rank weight sharing, candidate architectures reuse parameters while selected operations adapt to distinct input sources.

\subsection{Architecture Search}
\subsubsection{Search Space}

LRWS-NAO uses the ``middle'' cell search space from the implementation. Each architecture is encoded as two cells, a normal cell and a reduction cell. The normal cell preserves spatial resolution, while the reduction cell reduces spatial resolution by a factor of two when its edges originate from either of the two previous-cell inputs. Both cells are directed acyclic graphs with two input states, $s_{k-2}$ and $s_{k-1}$, and five intermediate nodes $v_2,\ldots,v_6$. Each intermediate node chooses two previous states and applies one operation to each chosen state:
\begin{equation}
  v_j = o_{j,1}(h_{i_{j,1}}) + o_{j,2}(h_{i_{j,2}}),
  \quad i_{j,1}, i_{j,2} < j,
\end{equation}
where $h_0=s_{k-2}$, $h_1=s_{k-1}$, and $h_j=v_j$ for intermediate nodes. The output of the cell is the channel-wise concatenation of all intermediate states that are not used as inputs by later nodes. Therefore, a cell can select not only its operations but also its effective output width through the number of unused states.

The final architecture string contains 40 integers: the first 20 define the normal cell and the last 20 define the reduction cell. Every four integers specify one intermediate node as
\begin{equation}
  (i_{j,1}, o_{j,1}, i_{j,2}, o_{j,2}).
\end{equation}
The two incoming edges are sampled independently, so they need not originate from different states. For example, the first node of the selected normal cell reads $s_{k-2}$ twice, passing it unchanged along one branch and applying a convolution along the other. To count the possible encodings, number the five intermediate nodes from $j=0$ to $B-1$ in order of construction. At node $j$, each edge has $j+2$ possible sources and $O$ possible operations. Since the two branches occupy separate slots in the sequence, the encoded cardinality of one cell is
\begin{equation}
  |\mathcal{C}_{B,O}| =
  \prod_{j=0}^{B-1} (j+2)^2 O^2
  = O^{2B}((B+1)!)^2.
\end{equation}
For the LRWS-NAO search space, $B=5$ and $O=10$, giving
\begin{equation}
  |\mathcal{C}_{5,10}| =
  10^{10}(2\cdot3\cdot4\cdot5\cdot6)^2
  = 5{,}184{,}000{,}000{,}000{,}000
  \approx 5.184\times10^{15}.
\end{equation}
Because a full architecture independently specifies one normal cell and one reduction cell, the encoded architecture-space size is
\begin{equation}
  |\mathcal{A}| = |\mathcal{C}_{5,10}|^2
  = (5.184\times10^{15})^2
  \approx 2.687\times10^{31}.
\end{equation}
Note that this calculation counts descriptions, not necessarily distinct computations. Exchanging the two complete branches of a sum produces a different sequence but leaves the sum unchanged; likewise, an edge with a zero operation can have more than one encoded source without changing its output.

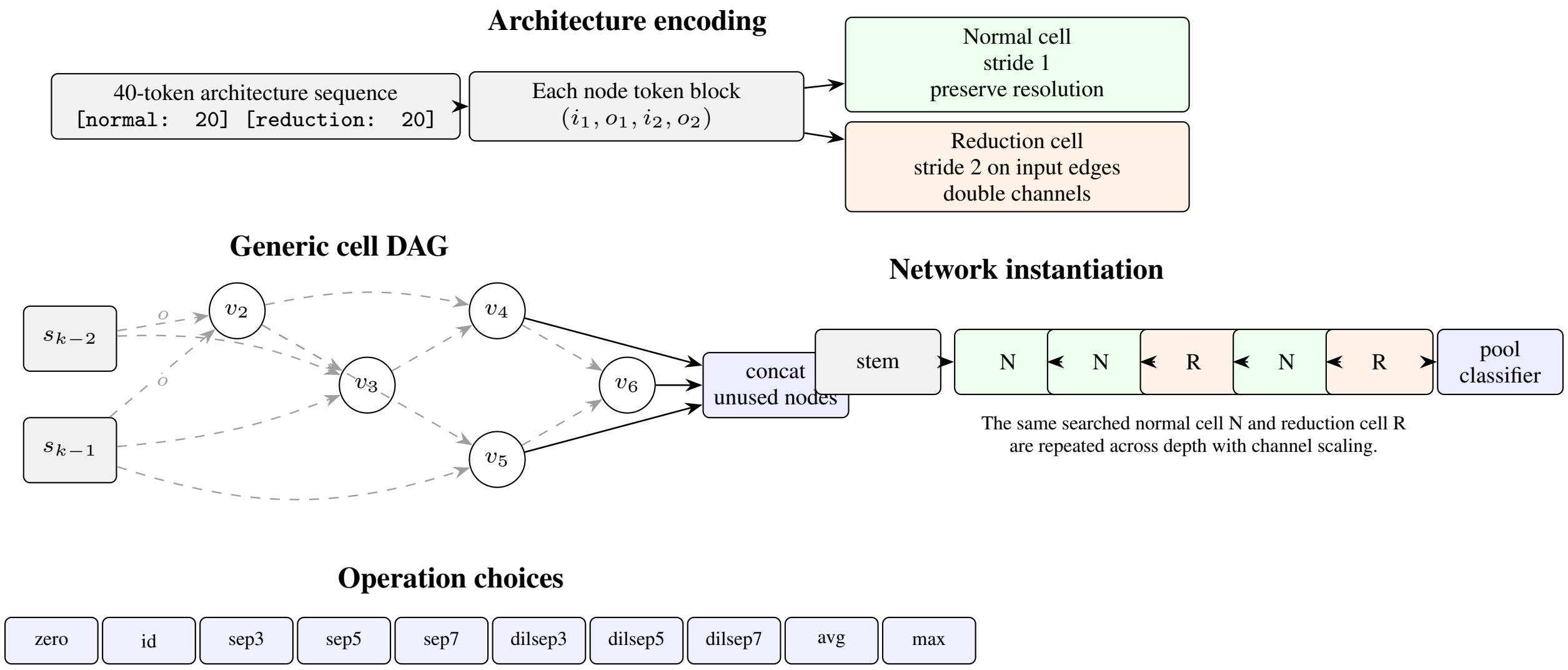
\begin{figure}[ht]
\centering
\resizebox{\textwidth}{!}{%
\begin{tikzpicture}[
  >=Stealth,
  block/.style={rectangle,rounded corners=2pt,draw,align=center,minimum height=7mm,font=\scriptsize,fill=white},
  codeblock/.style={rectangle,rounded corners=2pt,draw,align=center,minimum height=6mm,font=\scriptsize,fill=gray!10},
  cellnode/.style={circle,draw,minimum size=6mm,inner sep=1pt,font=\scriptsize,fill=white},
  opblock/.style={rectangle,rounded corners=2pt,draw,align=center,minimum height=5mm,font=\tiny,fill=blue!6},
  arrow/.style={->,line width=0.5pt},
  softarrow/.style={->,line width=0.45pt,dashed,gray!75},
  edgelabel/.style={font=\tiny,fill=white,inner sep=1pt}
]
\node[font=\small\bfseries] at (6.2,3.0) {Architecture encoding};
\node[codeblock,minimum width=4.4cm] (arch) at (2.2,2.1)
  {40-token architecture sequence\\
  \texttt{[normal: 20] [reduction: 20]}};
\node[codeblock,minimum width=3.6cm] (tuple) at (6.3,2.1)
  {Each node token block\\$(i_1,o_1,i_2,o_2)$};
\node[block,minimum width=3.7cm,fill=green!7] (normal) at (10.4,2.55)
  {Normal cell\\stride 1\\preserve resolution};
\node[block,minimum width=3.7cm,fill=orange!10] (reduc) at (10.4,1.45)
  {Reduction cell\\stride 2 on input edges\\double channels};
\draw[arrow] (arch) -- (tuple);
\draw[arrow] (tuple) -- (normal);
\draw[arrow] (tuple) -- (reduc);

\node[font=\small\bfseries] at (3.1,0.6) {Generic cell DAG};
\node[block,minimum width=1.0cm,fill=gray!12] (s0) at (0.2,-0.4) {$s_{k-2}$};
\node[block,minimum width=1.0cm,fill=gray!12] (s1) at (0.2,-1.6) {$s_{k-1}$};
\node[cellnode] (v2) at (2.0,-0.1) {$v_2$};
\node[cellnode] (v3) at (3.4,-0.9) {$v_3$};
\node[cellnode] (v4) at (4.8,-0.1) {$v_4$};
\node[cellnode] (v5) at (4.8,-1.7) {$v_5$};
\node[cellnode] (v6) at (6.2,-0.9) {$v_6$};
\node[block,minimum width=1.35cm,fill=blue!7] (concat) at (7.8,-0.9) {concat\\unused nodes};
\draw[softarrow] (s0) -- node[edgelabel,above] {$o$} (v2);
\draw[softarrow] (s1) -- node[edgelabel,below] {$o$} (v2);
\draw[softarrow] (s0) to[bend left=12] (v3);
\draw[softarrow] (s1) to[bend right=8] (v3);
\draw[softarrow] (v2) -- (v3);
\draw[softarrow] (v2) to[bend left=12] (v4);
\draw[softarrow] (v3) -- (v4);
\draw[softarrow] (s1) to[bend right=18] (v5);
\draw[softarrow] (v2) -- (v5);
\draw[softarrow] (v4) -- (v6);
\draw[softarrow] (v5) -- (v6);
\draw[arrow] (v4) -- (concat);
\draw[arrow] (v5) -- (concat);
\draw[arrow] (v6) -- (concat);

\node[font=\small\bfseries] at (10.5,0.35) {Network instantiation};
\node[block,minimum width=1.35cm,fill=gray!12] (stem) at (8.9,-0.65) {stem};
\node[block,minimum width=1.15cm,fill=green!7] (nA) at (10.3,-0.65) {N};
\node[block,minimum width=1.15cm,fill=green!7] (nB) at (11.3,-0.65) {N};
\node[block,minimum width=1.15cm,fill=orange!10] (rA) at (12.3,-0.65) {R};
\node[block,minimum width=1.15cm,fill=green!7] (nC) at (13.3,-0.65) {N};
\node[block,minimum width=1.15cm,fill=orange!10] (rB) at (14.3,-0.65) {R};
\node[block,minimum width=1.35cm,fill=blue!7] (cls) at (15.6,-0.65) {pool\\classifier};
\draw[arrow] (stem) -- (nA);
\draw[arrow] (nA) -- (nB);
\draw[arrow] (nB) -- (rA);
\draw[arrow] (rA) -- (nC);
\draw[arrow] (nC) -- (rB);
\draw[arrow] (rB) -- (cls);
\node[font=\tiny,align=center] at (12.3,-1.45)
  {The same searched normal cell N and reduction cell R\\are repeated across depth with channel scaling.};

\node[font=\small\bfseries] at (4.3,-3.0) {Operation choices};
\foreach \x/\name in {0/zero,1/id,2/sep3,3/sep5,4/sep7,5/dilsep3,6/dilsep5,7/dilsep7,8/avg,9/max} {
  \node[opblock,minimum width=1.0cm] at (\x*1.05,-3.65) {\name};
}
\end{tikzpicture}}
\caption{LRWS-NAO search space. Each architecture is a 40-token sequence split into normal-cell and reduction-cell encodings. Each intermediate node selects two previous states and two operations from the 10-operation middle search space. The classifier repeats the selected cells across its depth.}
\label{fig:searchspace}
\end{figure}

Figure~\ref{fig:searchspace} summarizes the choices available at each node. Table~\ref{tab:cardinality} gives the number of possible encodings, and Table~\ref{tab:ops} maps the operation IDs to their corresponding operations.

\begin{table}[ht]
\centering
\caption{Encoded search-space cardinality for the LRWS-NAO middle search space.}
\label{tab:cardinality}
\begin{tabular}{lll}
\toprule
Quantity & Formula & Value for $B=5,O=10$ \\
\midrule
Choices at node $j$ & $(j+2)^2O^2$ & $j=0,\ldots,4$ \\
One cell & $O^{2B}((B+1)!)^2$ & $5.184\times10^{15}$ \\
Normal + reduction pair & $|\mathcal{C}_{B,O}|^2$ & $2.687\times10^{31}$ \\
Initial random seed pool & $N_0$ & $2000$ architectures \\
Seed-pool fraction & $N_0/|\mathcal{A}|$ & $\approx 7.44\times10^{-29}$ \\
\bottomrule
\end{tabular}
\end{table}

\begin{table}[ht]
\centering
\caption{Operation IDs in the LRWS-NAO middle search space. In reduction cells, identity on an input edge is implemented as factorized reduction because the stride is two.}
\label{tab:ops}
\begin{tabular}{clcl}
\toprule
ID & Operation & ID & Operation \\
\midrule
0 & Zero & 5 & Dilated separable convolution 3x3 \\
1 & Identity / factorized reduction & 6 & Dilated separable convolution 5x5 \\
2 & Separable convolution 3x3 & 7 & Dilated separable convolution 7x7 \\
3 & Separable convolution 5x5 & 8 & Average pooling 3x3 \\
4 & Separable convolution 7x7 & 9 & Max pooling 3x3 \\
\bottomrule
\end{tabular}
\end{table}

\subsubsection{Low-Rank Weight Sharing in the Supernet}

During search we train a single network that contains the operations needed by many candidate cells. Each minibatch activates only the edges of one sampled architecture, so a candidate can later be scored with weights that have already received training updates. To keep these evaluations inexpensive, the proxy uses 24 initial channels and a depth setting of 3, which expands to 11 cells across the stages. The proxy score is a basis for comparison among candidates, not a forecast of the accuracy of the wider classifier.

\begin{figure}[ht]
\centering
\resizebox{\textwidth}{!}{%
\begin{tikzpicture}[
  >=Stealth,
  panel/.style={rectangle,rounded corners=3pt,draw=black!45,minimum width=7.6cm,minimum height=5.2cm,fill=white},
  block/.style={rectangle,rounded corners=2pt,draw,align=center,minimum height=7mm,font=\scriptsize,fill=white},
  small/.style={rectangle,rounded corners=2pt,draw,align=center,minimum height=5mm,font=\tiny,fill=white},
  tensor/.style={rectangle,rounded corners=2pt,draw,align=center,minimum width=1.25cm,minimum height=6mm,font=\scriptsize,fill=green!10},
  adapter/.style={rectangle,rounded corners=2pt,draw,align=center,minimum width=1.15cm,minimum height=6mm,font=\scriptsize,fill=orange!14},
  outblock/.style={rectangle,rounded corners=2pt,draw,align=center,minimum width=1.25cm,minimum height=6mm,font=\scriptsize,fill=blue!8},
  arrow/.style={->,line width=0.55pt},
  thin/.style={->,line width=0.4pt},
  dashedarrow/.style={->,line width=0.45pt,dashed,gray!70},
  lab/.style={font=\tiny,fill=white,inner sep=1pt}
]

\node[panel] (leftpanel) at (3.8,0) {};
\node[panel] (rightpanel) at (12.4,0) {};
\node[font=\small\bfseries] at (3.8,2.25) {(a) Conventional weight sharing};
\node[font=\small\bfseries] at (12.4,2.25) {(b) Low-rank adapted weight sharing};

\node[small,fill=gray!10] (la) at (0.9,1.15) {edge A\\$(x_a,o)$};
\node[small,fill=gray!10] (lb) at (0.9,0.15) {edge B\\$(x_b,o)$};
\node[small,fill=gray!10] (lc) at (0.9,-0.85) {edge C\\$(x_c,o)$};
\node[tensor] (lw) at (3.35,0.15) {shared\\$W_o$};
\node[block,fill=yellow!10,minimum width=1.7cm] (lop) at (5.35,0.15) {$o(x;W_o)$};
\node[outblock] (lo) at (6.75,0.15) {output};
\draw[dashedarrow] (la) -- (lw);
\draw[dashedarrow] (lb) -- (lw);
\draw[dashedarrow] (lc) -- (lw);
\draw[arrow] (lw) -- (lop);
\draw[arrow] (lop) -- (lo);
\node[font=\tiny,align=center] at (3.8,-1.85)
  {All sampled subgraphs update the same operation weight.\\
  This is cheap, but different graph contexts can interfere.};

\node[small,fill=gray!10] (rx) at (8.95,0.9) {selected edge\\$(x_i,o)$};
\node[tensor] (rw) at (10.75,1.15) {base\\$W_o$};
\node[adapter] (ra) at (10.25,-0.25) {$A_i^o$};
\node[adapter] (rb) at (11.35,-0.25) {$B_i^o$};
\node[block,fill=orange!8,minimum width=1.8cm] (delta) at (12.75,-0.25) {$\Delta W_i^o=A_i^oB_i^o$};
\node[block,fill=yellow!12,minimum width=1.85cm] (sum) at (13.25,1.15) {$\widetilde{W}_{i,o}=W_o+\Delta W_i^o$};
\node[block,fill=blue!7,minimum width=1.65cm] (rop) at (15.1,1.15) {$o(x_i;\widetilde{W}_{i,o})$};
\node[outblock] (ro) at (16.55,1.15) {output};
\draw[arrow] (rx) -- (rw);
\draw[thin] (rx) to[bend right=20] (ra);
\draw[arrow] (ra) -- (rb);
\draw[arrow] (rb) -- (delta);
\draw[arrow] (delta) -- (sum);
\draw[arrow] (rw) -- (sum);
\draw[arrow] (sum) -- (rop);
\draw[arrow] (rop) -- (ro);
\node[font=\tiny,align=center] at (12.65,-1.85)
  {The base operation remains shared.\\
  A small rank-$r$ correction specializes the selected input/operation.\\
  In the search run, $r=16$ and adapter weights use a higher learning rate.};

\node[font=\tiny,align=center] at (8.1,-2.65)
  {LRWS-NAO uses this adaptation only in the search supernet; the final decoded cell is trained from scratch.};
\end{tikzpicture}}
\caption{Low-rank adapted weight sharing used during LRWS-NAO search. Conventional one-shot NAS uses one shared operation weight $W_o$ for many sampled edges. LRWS keeps the shared base weight but adds an input- and operation-specific low-rank correction $\Delta W_i^o=A_i^oB_i^o$, producing the adapted weight $\widetilde{W}_{i,o}=W_o+\Delta W_i^o$.}
\label{fig:lora_ws}
\end{figure}
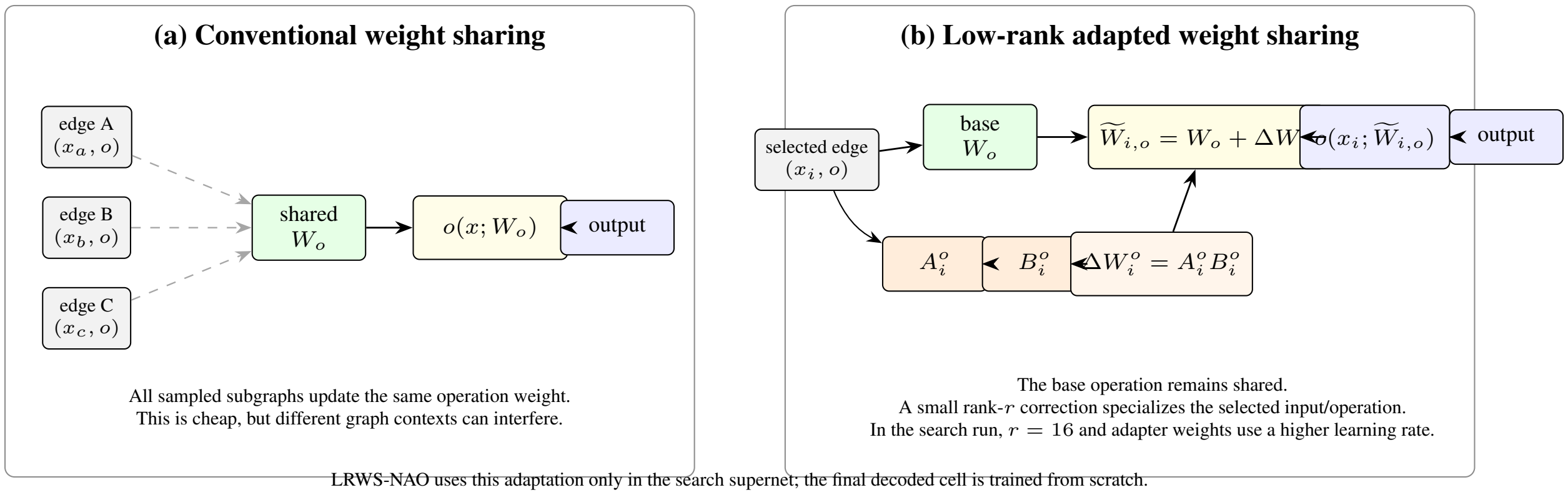

Consider the same convolution applied first to a feature map from the preceding cell and then to a feature map produced within the current cell. Under ordinary weight sharing, an update prompted by the first input also changes the filter used for the second. Separate filters would avoid this coupling but would discard the benefit of sharing. Instead, the low-rank adapter \cite{hu2022lora} gives each source $i$ a correction to the same base kernel $W$:
\begin{equation}
  W_i' = W + A_i B_i,
\end{equation}
where $A_i$ and $B_i$ are low-rank adapter matrices for the $i$-th possible input. We apply low-rank adapters to shared separable and dilated separable convolutions. For separable convolution, adapters are attached to both depthwise and pointwise components. The adapter $B$ matrices are initialized to zero, so the adapted operation initially behaves like the base shared operation; the adapter $A$ matrices are initialized with Kaiming initialization. This makes the search begin from the original weight-sharing behavior and then gradually specialize as training proceeds. Figure~\ref{fig:lora_ws} illustrates the mechanism.

The base kernels and their adapters are trained together, with a rank of 16 and a learning rate ten times the base rate for the classification adapters. This gives each source some freedom to depart from the common filter during search. Once a cell has been selected, only its connections and operations are retained: the final classifier uses ordinary convolutions initialized from scratch, so its parameter count includes no search adapters.

\subsubsection{Hybrid NAO and PPO/CVaR Generation}

The original NAO loop trains an encoder-predictor-decoder model over evaluated architectures, then improves top architectures through gradient steps in the learned embedding space. Let $a$ be a discrete architecture sequence. The encoder maps $a$ to a continuous embedding $z=E(a)$; the predictor estimates performance $\hat{y}=f(z)$; and the decoder maps an improved embedding back to a valid architecture $\hat{a}=D(z')$. The standard NAO update moves the embedding of a promising architecture in a direction that increases predicted performance:
\begin{equation}
  z' = z + \eta \nabla_z f(z),
\end{equation}
followed by decoding and duplicate filtering. This keeps the search informed by previous evaluations rather than sampling blindly from the full discrete space.

After the initial proposal round, LRWS-NAO adds a second way to move from a promising architecture. Rather than always following the predictor gradient, a PPO policy \cite{schulman2017ppo} learns a distribution of moves from their predicted gains. Its input state $s=[z,\hat{y}]$ combines the architecture embedding $z$ with the predicted accuracy $\hat{y}=f_{\phi}(z)$. The actor is a Gaussian policy
\begin{equation}
  u \sim \pi_{\psi}(u\mid s)
  = \mathcal{N}\left(\mu_{\psi}(s), \operatorname{diag}(\sigma_{\psi}^2)\right),
\end{equation}
where $u$ is a continuous step direction in the NAO embedding space. To prevent unstable jumps, the implementation clips the step norm:
\begin{equation}
  \bar{u} = u \cdot \min\left(1,\frac{m}{\|u\|_2}\right),
  \quad m=3.0.
\end{equation}
The moved encoder representation is normalized and decoded by the NAO decoder:
\begin{equation}
  z' = \operatorname{norm}(z+\bar{u}), \qquad
  a' = D_{\phi}(z').
\end{equation}
In this notation $z'$ summarizes the moved representation. In the implementation, the same action vector is added to each encoder position, the new vectors are normalized, and their normalized mean supplies the new embedding. During rollouts the actor samples a move to learn which directions receive higher rewards; when generating candidates for the pool, it uses the policy mean and passes the moved representation to the NAO decoder. PPO therefore chooses a direction, while NAO determines which discrete architecture that direction produces.

Evaluating a new classifier for every sampled move would defeat the purpose of searching in a continuous representation. Instead, the actor receives the change in the predictor's normalized score before the proposed cell is evaluated with the supernet:
\begin{equation}
  r = f_{\phi}(z') - f_{\phi}(z).
\end{equation}
For example, two moves that end at the same predicted score receive different rewards if one began at a lower score. This rewards improvement over the starting architecture rather than simply rewarding a favorable starting point. The differences are standardized within each rollout minibatch; they are neither measured accuracy gains nor gains in percentage points.

To compare a move with what the critic expected, the trainer computes generalized advantage estimates with $\gamma=0.99$ and $\lambda=0.95$. Here $t$ indexes the order of samples in the rollout buffer: starting architectures are sampled from the pool, so adjacent entries need not be successive edits to the same architecture. With critic value $V_{\psi}$, the recursion uses
\begin{align}
  \delta_t &= r_t + \gamma V_{\psi}(s_{t+1}) - V_{\psi}(s_t),\\
  \hat{A}_t &= \sum_{\ell=0}^{\infty}(\gamma\lambda)^{\ell}\delta_{t+\ell}.
\end{align}
After standardizing the advantages, PPO compares the probability density of the same sampled action under the updated policy and the policy that generated it. A ratio above one means that the updated policy gives that move greater density:
\begin{equation}
  \rho_t(\psi) =
  \frac{\pi_{\psi}(u_t\mid s_t)}
       {\pi_{\psi_{\mathrm{old}}}(u_t\mid s_t)},
\end{equation}
and the clipped surrogate for each sample is
\begin{equation}
  \ell_t(\psi)=
  \min\left(
    \rho_t(\psi)\hat{A}_t,\,
    \operatorname{clip}(\rho_t(\psi),1-\epsilon,1+\epsilon)\hat{A}_t
  \right),
  \quad \epsilon=0.2.
\end{equation}

Clipping limits the incentive to change the probability of a sampled move too far in one update. To decide which moves affect the classification actor, we first rank the $n$ entries of each PPO minibatch by their \emph{unclipped} scores $\rho_t\hat{A}_t$. We then retain the lowest-scoring $k=\max(1,\lfloor\alpha n\rfloor)$ entries,
\begin{equation}
  I_{\alpha} =
  \operatorname{indices\_of\_smallest}_{k}
  \{\rho_t\hat{A}_t\}_{t=1}^{n},
  \quad \alpha=0.2.
\end{equation}
The actor objective then maximizes the clipped surrogate on this lower-tail subset:
\begin{equation}
  J_{\mathrm{CVaR-PPO}}(\psi)
  = \frac{1}{|I_{\alpha}|}
    \sum_{t\in I_{\alpha}}\ell_t(\psi).
\end{equation}
The minimized loss is
\begin{equation}
  \mathcal{L}(\psi)
  = -J_{\mathrm{CVaR-PPO}}(\psi)
    + c_v\,\frac{1}{n}\sum_t\left(V_{\psi}(s_t)-\hat{R}_t\right)^2
    - c_e\,\frac{1}{n}\sum_t \mathcal{H}\!\left[\pi_{\psi}(\cdot\mid s_t)\right],
\end{equation}
with $c_v=0.5$ and $c_e=0.01$ in the reported run. For a minibatch of 64 moves, the actor term averages the clipped scores of 12 moves, whereas the critic and entropy terms still use the full minibatch. This distinction matters: the tail term gives priority to weak policy moves, rather than to images the classifier finds difficult.

The search starts from 2,000 architectures and trains the supernet for 80 epochs in each of four iterations. After each of the first three iterations, the NAO controller is fitted for 1,000 epochs with a target of 500 proposals for the next pool. The first round uses predictor-gradient proposals alone; the next two supplement them with PPO proposals. The fourth iteration evaluates the accumulated pool, after which the five highest-scoring entries are trained independently for 50 epochs to select the architecture for full training.

\subsubsection{Algorithms}

The algorithms below distinguish the three uses of a score. Algorithm~\ref{alg:sample} defines the legal encodings, and Algorithm~\ref{alg:search} evaluates those encodings with shared weights. Within that search, Algorithm~\ref{alg:ppo} uses predicted gains to propose further candidates. Only after independent reranking does Algorithm~\ref{alg:train} train the selected topology at its final width.

\begin{algorithm}[H]
\caption{Sampling an Encoded Cell and Architecture}
\label{alg:sample}
\begin{algorithmic}[1]
\Require Number of intermediate nodes $B$, number of operations $O$
\Ensure Architecture $a=(c_{\mathrm{normal}}, c_{\mathrm{reduction}})$
\Function{SampleCell}{$B,O$}
  \State $c \gets [\,]$
  \For{$j=0$ to $B-1$}
    \State $p_1 \sim \mathrm{Uniform}\{0,\ldots,j+1\}$
    \State $o_1 \sim \mathrm{Uniform}\{0,\ldots,O-1\}$
    \State $p_2 \sim \mathrm{Uniform}\{0,\ldots,j+1\}$
    \State $o_2 \sim \mathrm{Uniform}\{0,\ldots,O-1\}$
    \State Append $(p_1,o_1,p_2,o_2)$ to $c$
  \EndFor
  \State \Return $c$
\EndFunction
\State $c_{\mathrm{normal}} \gets \Call{SampleCell}{B,O}$
\State $c_{\mathrm{reduction}} \gets \Call{SampleCell}{B,O}$
\State \Return $(c_{\mathrm{normal}}, c_{\mathrm{reduction}})$
\end{algorithmic}
\end{algorithm}

\begin{pagedalgorithm}{Full LRWS-NAO Architecture Search}
\small
\label{alg:search}
\begin{algorithmic}[1]
\Require Search data, validation data, seed size $N_0=2000$, search iterations $T=4$, child epochs $E=80$, new-architecture budget $K=500$
\Ensure Best discovered architecture $a^\star$
\State Initialize $\mathcal{P}_0$ with $N_0$ valid random architectures using Algorithm~\ref{alg:sample}
\State Initialize search supernet $S_{\theta}$ with shared weights $W$ and low-rank adapters $\{A,B\}$
\State Split optimizer parameters into base group $W$ and adapter group $\{A,B\}$ with LR multiplier 10
\State Initialize NAO controller $G_{\phi}=(E_{\phi},f_{\phi},D_{\phi})$
\State Initialize Gaussian actor-critic policy $\pi_{\psi}$ for latent moves
\State $\mathcal{H}\gets\emptyset$ \Comment{archive of evaluated architectures}
\For{$t=0$ to $T-1$}
  \State Reset or continue the child optimizer and learning-rate schedule for the current pool
  \For{$e=1$ to $E$}
    \For{each search-training minibatch $(x,y)$}
      \State Sample architecture $a\sim\mathcal{P}_t$
      \State Decode $a$ into normal and reduction cell DAGs
      \State Activate the corresponding subgraph in $S_{\theta}$
      \State For every selected convolutional edge, use $\widetilde{W}_{i,o}=W_o+A_i^oB_i^o$
      \State Compute label-smoothed cross-entropy loss on $(x,y)$
      \State Backpropagate through only the active subgraph and update $W,A,B$
    \EndFor
  \EndFor
  \State Append the current pool $\mathcal{P}_t$ to archive $\mathcal{H}$
  \For{each architecture entry $a$ in $\mathcal{H}$}
    \State Recompute its validation score with the current supernet weights
  \EndFor
  \State Sort $\mathcal{H}$ by validation accuracy
  \If{$t=T-1$}
    \State \textbf{break}
  \EndIf
  \State Encode each architecture as a 40-token sequence and normalize scores to $[0,1]$
  \For{$c=1$ to 1000}
    \State Update $G_{\phi}$ using reconstruction loss, performance-prediction loss, and length-valid decoding
  \EndFor
  \State Let $\mathcal{Q}$ be the top 100 architectures in $\mathcal{H}$
  \State $\mathcal{C}\gets\emptyset$ \Comment{new candidate set}
  \If{$t=0$}
    \State Generate candidates by NAO latent updates $z'=z+\eta\nabla_z f_{\phi}(z)$ from $\mathcal{Q}$
    \State Add all decoded valid, nonduplicate NAO candidates to $\mathcal{C}$
  \Else
    \State Generate roughly $K/2$ candidates by NAO latent-gradient decoding and add valid candidates to $\mathcal{C}$
    \State Call Algorithm~\ref{alg:ppo} with $\mathcal{Q}$, $G_{\phi}$, and $\pi_{\psi}$ to update the policy
    \State Decode policy-mean latent moves from $\mathcal{Q}$ and add valid, nonduplicate candidates to $\mathcal{C}$
    \If{$|\mathcal{C}|<K$}
      \State Fill the remaining budget with additional NAO-gradient candidates
    \EndIf
  \EndIf
  \State Add noise-perturbed variants of high-ranking architectures when more diversity is needed
  \State Filter $\mathcal{C}$ by sequence length, node-index bounds, operation-token bounds, and duplicate checks
  \State $\mathcal{P}_{t+1}\gets$ decoded valid architectures from $\mathcal{C}$
\EndFor
\State Rerank the top 5 architectures from $\mathcal{H}$ by short train-and-evaluate runs
\State \Return best reranked architecture $a^\star$
\end{algorithmic}
\end{pagedalgorithm}

\begin{pagedalgorithm}{PPO/CVaR Latent Architecture Generator}
\small
\label{alg:ppo}
\begin{algorithmic}[1]
\Require Top architecture set $\mathcal{Q}$, NAO controller $G_{\phi}=(E_{\phi},f_{\phi},D_{\phi})$, policy $\pi_{\psi}$, rollout size $R=2048$, CVaR level $\alpha=0.2$
\Ensure Updated policy $\pi_{\psi}$ and policy-generated candidate architectures
\State $\mathcal{D}_{\mathrm{ppo}}\gets\emptyset$
\For{$r=1$ to $R$}
  \State Sample architecture $a$ from $\mathcal{Q}$
  \State Encode $z=E_{\phi}(a)$ and predict $\hat{y}=f_{\phi}(z)$
  \State Form policy state $s=[z,\hat{y}]$
  \State Sample latent action $u\sim\mathcal{N}(\mu_{\psi}(s),\operatorname{diag}(\sigma_{\psi}^{2}))$
  \State Store old log probability $\log\pi_{\psi_{\mathrm{old}}}(u\mid s)$ and critic value $V_{\psi}(s)$
  \State Clip action norm $\bar{u}=u\min(1,3/\|u\|_2)$
  \State Move and normalize latent code $z'=\operatorname{norm}(z+\bar{u})$
  \State Compute surrogate reward $r_{\mathrm{ppo}}=f_{\phi}(z')-f_{\phi}(z)$
  \State Add transition $(s,u,r_{\mathrm{ppo}},V_{\psi}(s),\log\pi_{\psi_{\mathrm{old}}}(u\mid s),z')$ to $\mathcal{D}_{\mathrm{ppo}}$
\EndFor
\State Standardize rewards within each rollout minibatch
\State Compute GAE advantages $\hat{A}_t$ using $\gamma=0.99$ and $\lambda=0.95$
\State Form returns from the original advantages and critic values; standardize advantages for the actor
\For{each PPO epoch and minibatch}
  \State Recompute $\log\pi_{\psi}(u_t\mid s_t)$, entropy $\mathcal{H}_t$, and value $V_{\psi}(s_t)$
  \State $\rho_t=\exp(\log\pi_{\psi}(u_t\mid s_t)-\log\pi_{\psi_{\mathrm{old}}}(u_t\mid s_t))$
  \State $\ell_t=\min(\rho_t\hat{A}_t,\operatorname{clip}(\rho_t,1-\epsilon,1+\epsilon)\hat{A}_t)$ with $\epsilon=0.2$
  \State Select the $k=\max(1,\lfloor\alpha n\rfloor)$ lowest values of $\rho_t\hat{A}_t$ to form $I_{\alpha}$
  \State $J_{\mathrm{CVaR}}=|I_{\alpha}|^{-1}\sum_{t\in I_{\alpha}}\ell_t$
  \State $\mathcal{L}=-J_{\mathrm{CVaR}}+0.5\,\mathrm{MSE}(V_{\psi}(s_t),\hat{R}_t)-0.01\,\mathrm{mean}(\mathcal{H}_t)$
  \State Update $\psi$ by gradient descent on $\mathcal{L}$
\EndFor
\For{each architecture $a\in\mathcal{Q}$}
  \State Use policy mean $\mu_{\psi}([E_{\phi}(a),f_{\phi}(E_{\phi}(a))])$ as a deterministic latent move
  \State Decode $D_{\phi}(\operatorname{norm}(E_{\phi}(a)+\bar{u}))$ and keep valid nonduplicates
\EndFor
\State \Return updated $\pi_{\psi}$ and generated candidates
\end{algorithmic}
\end{pagedalgorithm}

\begin{algorithm}[H]
\small
\caption{Stand-Alone Training of the Selected Classifier}
\label{alg:train}
\begin{algorithmic}[1]
\Require Selected architecture $a^\star$, epochs $1000$, channels $192$, layers $8$
\Ensure Best checkpoint and held-out classification accuracy
\State Build the classifier from the discrete normal and reduction cells in $a^\star$
\State Enable cutout, AutoAugment, label smoothing, auxiliary head, drop-path, and SWA
\State Initialize SGD with cosine learning-rate decay and weight decay $5\times10^{-4}$
\State $b\gets 0$ \Comment{best evaluation accuracy}
\For{$e=0$ to $999$}
  \For{each training minibatch $(x,y)$}
    \State Compute logits and auxiliary logits
    \State $\mathcal{L}\gets \mathcal{L}_{\mathrm{CE/LS}}(\mathrm{logits},y) + 0.4\,\mathcal{L}_{\mathrm{CE/LS}}(\mathrm{aux},y)$
    \State Update model parameters with SGD
  \EndFor
  \State Measure classification accuracy on the held-out split
  \If{accuracy $> b$}
    \State Save best checkpoint and update $b$
  \EndIf
  \If{SWA is enabled and $e\ge750$}
    \State Update the averaged SWA model
  \EndIf
\EndFor
\State Recompute SWA batch-normalization statistics and evaluate SWA model
\State \Return best non-SWA checkpoint and SWA result
\end{algorithmic}
\end{algorithm}

\FloatBarrier
\subsection{Theoretical Analysis}

The analysis separates three quantities that are easy to confuse: the chance of sampling an encoding, the gain predicted for a proposed change, and the accuracy obtained after independent training. The first can be calculated from the search space, whereas connecting a predicted gain to a real gain requires assumptions about prediction error. We make those assumptions explicit before interpreting the effect of low-rank adaptation and the classification policy objective.

\subsubsection{Search-Space Coverage and Random Acceptance Probability}

Let $M=|\mathcal{A}|$ be the encoded normal-plus-reduction search-space size. For LRWS-NAO, $M\approx2.687\times10^{31}$. If $N$ architectures are sampled independently and uniformly from this space, the probability of sampling any fixed target architecture is
\begin{equation}
  P_{\mathrm{hit}}(N)
  = 1-\left(1-\frac{1}{M}\right)^N
  \approx \frac{N}{M}
  \quad \text{for } N\ll M.
\end{equation}
For the initial seed pool $N=2000$, this gives
\begin{equation}
  P_{\mathrm{hit}}(2000) \approx 7.44\times10^{-29}.
\end{equation}
Thus, the initial pool is useful because it provides examples from which the supernet and predictor can learn, even though its probability of containing any one prespecified architecture is negligible.

More generally, suppose a fraction $p_{\tau}$ of architectures reaches an accuracy threshold $\tau$ under full training. The probability that uniform random search finds at least one such architecture after $N$ samples is
\begin{equation}
  P_{\tau}(N)=1-(1-p_{\tau})^N.
\end{equation}
Equivalently, to obtain confidence $1-\delta$, the required number of random samples is
\begin{equation}
  N \geq \frac{\log \delta}{\log(1-p_{\tau})}
  \approx \frac{\log(1/\delta)}{p_{\tau}}
  \quad \text{for small } p_{\tau}.
\end{equation}
If high-performing architectures occupy a very small fraction of the space, random search rapidly becomes infeasible even when collisions are essentially absent. Indeed, the expected number of duplicated samples in $N$ uniform draws is approximately
\begin{equation}
  \mathbb{E}[\mathrm{collisions}]
  \approx \frac{N(N-1)}{2M}.
\end{equation}
For $N=2000$, the expected number of colliding pairs is about $7.44\times10^{-26}$. Nearly every sample can therefore be different even when the pool covers only a negligible fraction of the space.

A proposal generated by the controller must pass additional checks. Let $q$ be its probability of having a valid encoding, $d$ its probability of being new given validity, and $p_{\tau}^{\mathrm{prop}}$ its probability of exceeding threshold $\tau$ given both conditions. Their product is the probability of a new, valid proposal above the threshold. Under the simplifying assumption of $K$ independent proposals with the same probabilities, the chance of at least one such proposal is
\begin{equation}
  P_{\mathrm{accept}}(K,\tau)
  = 1-\left(1-q\,d\,p_{\tau}^{\mathrm{prop}}\right)^K.
\end{equation}
This expression is a reference calculation rather than an empirical acceptance rate: proposals in the actual search share a predictor, and the set of previously seen architectures changes after a proposal is accepted.

\subsubsection{Predictor-Guided Improvement and Error Sensitivity}

Consider two architectures $a$ and $a'$ with unknown stand-alone validation accuracies $y(a)$ and $y(a')$. For this argument, suppose that predictor scores $\hat{y}$ are expressed on the same scale as those accuracies. Write the predicted gain as
\begin{equation}
  \Delta_{\mathrm{pred}} = \hat{y}(a')-\hat{y}(a).
\end{equation}
If the predictor error is uniformly bounded on the two architectures,
\begin{equation}
  |\hat{y}(a)-y(a)|\leq \epsilon,
  \qquad
  |\hat{y}(a')-y(a')|\leq \epsilon,
\end{equation}
then the true gain satisfies
\begin{equation}
  y(a')-y(a)
  \geq \Delta_{\mathrm{pred}} - 2\epsilon.
\end{equation}
The factor of two matters because the predictor may underestimate the starting architecture by $\epsilon$ while overestimating its replacement by the same amount. Only a predicted gain larger than $2\epsilon$ guarantees the correct direction under this assumption. We do not know such an error bound for the fitted predictor, which learns normalized supernet scores rather than stand-alone accuracies. Independent reranking is therefore an empirical check on the ordering, not an application of a certified threshold.

A further distinction arises during classification rollouts: the policy receives a reward at the moved embedding, before evaluating the decoded architecture. In the notation of the controller, this is
\begin{equation}
  r=f_{\phi}(z')-f_{\phi}(z).
\end{equation}
To interpret this reward as an accuracy gain, the predictor would need to be calibrated and the decoded architecture would need to retain the advantage predicted for its embedding. Neither property follows from using PPO.

\subsubsection{Low-Rank Adaptation as Controlled Supernet Specialization}

Plain weight sharing can be viewed as imposing a single parameter vector $W_o$ for operation $o$ across many graph contexts. If context $c$ denotes an edge location, input source, and surrounding cell structure, the best context-specific operation weight would generally be $W_{o,c}^{\star}$. Plain sharing constrains every context to use the same approximation:
\begin{equation}
  W_{o,c}\equiv W_o.
\end{equation}
This can create a bias term in the shared-weight validation estimate because some architectures are evaluated with operation weights that are better suited to other architectures.

LRWS-NAO relaxes this constraint by using a low-rank correction,
\begin{equation}
  W_{o,c}=W_o+A_{o,c}B_{o,c},
  \qquad \operatorname{rank}(A_{o,c}B_{o,c})\leq r.
\end{equation}
For a flattened convolutional weight matrix $W\in\mathbb{R}^{d_{\mathrm{out}}\times d_{\mathrm{in}}}$, a full context-specific copy would require $d_{\mathrm{out}}d_{\mathrm{in}}$ additional parameters per context. A rank-$r$ adapter requires only
\begin{equation}
  r(d_{\mathrm{out}}+d_{\mathrm{in}})
\end{equation}
parameters, giving an adapter-to-full-copy ratio of
\begin{equation}
  \frac{r(d_{\mathrm{out}}+d_{\mathrm{in}})}
       {d_{\mathrm{out}}d_{\mathrm{in}}}.
\end{equation}
The parameter ratio is below one when $r(d_{\mathrm{out}}+d_{\mathrm{in}})<d_{\mathrm{out}}d_{\mathrm{in}}$, so the saving depends on kernel dimensions as well as rank. Because one factor starts at zero, each adapted kernel initially equals its shared base. Subsequent updates can specialize it for a source, but only within the rank constraint: note that $r$ bounds the number of independent directions in the correction, not the magnitude of its entries.

\begin{figure}[tbp]
\centering
\includegraphics[width=\linewidth]{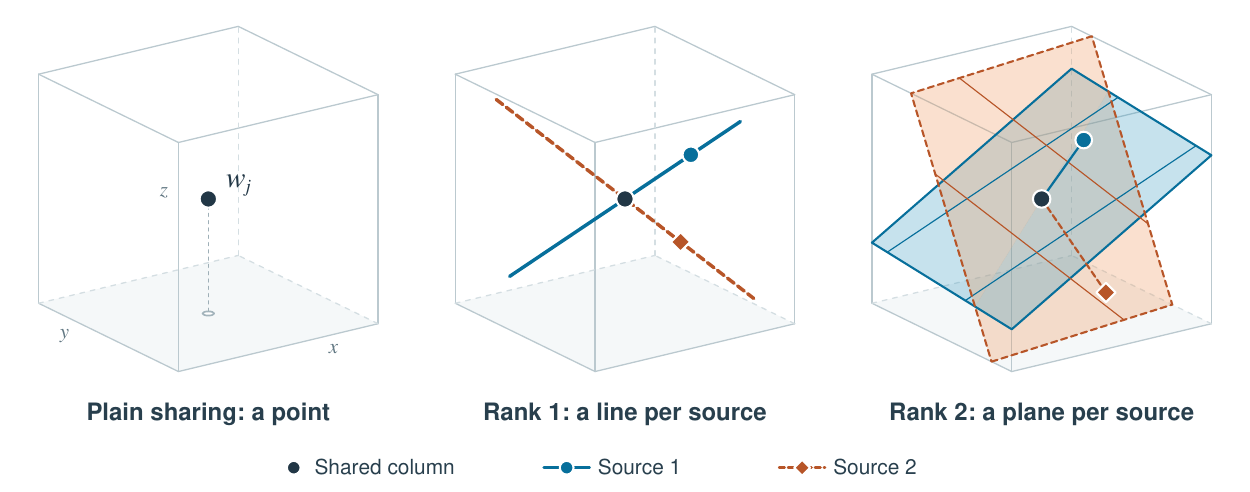}
\caption{The same shared column $w_j$ (dark point) is shown in three output-channel coordinates. The blue circle and orange diamond mark example weights for two input sources. With one learned direction per source, each example lies on its source's line through $w_j$; with two independent directions, it can lie anywhere on its source's plane. The cube marks the viewing window, not a bound on the correction.}
\label{fig:weight_sharing_geometry}
\end{figure}

To see what rank controls, consider one column $w_j$ of a shared kernel with three output channels, so that $w_j$ is a point in $\mathbb{R}^3$. Under plain sharing, both input sources use that same point. For a rank-one correction $\Delta W_i=u_i v_i^\top$, the adapted column is $w_j+v_{i,j}u_i$: once the source direction $u_i$ is specified, changing the coefficient $v_{i,j}$ moves the column along a line through $w_j$. A second independent direction expands that line to a plane, as Figure~\ref{fig:weight_sharing_geometry} illustrates. The base and the directions are learned during training; we hold them fixed in the illustration to isolate the effect of changing the coefficients. More generally, at rank at most $r$, it is the \emph{displacements} of a source's columns, not the adapted columns themselves, that span at most $r$ dimensions.

The adapters are discarded after selection, so their purpose is to change how candidates are scored during search rather than to increase the capacity of the final classifier. To see what would have to improve, consider the following idealized decomposition of a shared-weight score:
\begin{equation}
  \hat{y}_{\mathrm{ws}}(a)=y(a)+b_{\mathrm{share}}(a)+\xi(a),
\end{equation}
where $b_{\mathrm{share}}(a)$ is architecture-dependent bias induced by shared weights and $\xi(a)$ is stochastic evaluation noise. Low-rank adaptation is useful when it reduces the variance or magnitude of $b_{\mathrm{share}}(a)$ enough to improve ranking, while not adding so much context capacity that the supernet overfits the sampled architecture pool.

\subsubsection{CVaR-PPO as Lower-Tail Control}

Consider a policy that proposes a few highly rated moves and many slightly unfavorable ones. Its average score can rise even if its weaker moves do not improve. The classification actor instead uses the lower portion of a minibatch of move scores, which changes the observations that contribute directly to the policy gradient.

More precisely, let $U_t=\rho_t\hat{A}_t$ be the unclipped surrogate and $X_t=\ell_t(\psi)$ its clipped counterpart. If $I_\alpha$ contains the $k=\max(1,\lfloor\alpha n\rfloor)$ smallest values of $U_t$, then the implemented objective is
\begin{equation}
  J_{\mathrm{tail}}
  =
  \frac{1}{k}
  \sum_{t\in I_{\alpha}} X_t,
\end{equation}
where selection and averaging use different scores. Since $X_t\leq U_t$ for each sample and the mean of the smallest $k$ values of $U_t$ cannot exceed its full mean,
\begin{equation}
  J_{\mathrm{tail}} \leq \frac{1}{k}\sum_{t\in I_\alpha}U_t
  \leq \frac{1}{n}\sum_{t=1}^{n}U_t.
\end{equation}
This bound is with respect to the \emph{unclipped} mean. Selecting samples by $U_t$ need not give the same subset as selecting by $X_t$, so the objective is inspired by lower-tail CVaR rather than being the exact tail mean of the clipped scores.

The step constraint and the tail criterion consequently play different roles: the constraint $\|\bar{u}\|_2\leq3$ limits how far the policy can move, while the tail criterion determines which sampled moves influence its update. Neither imposes a lower bound on the accuracy of a decoded cell or on performance for any subgroup of images.

\subsubsection{Parameter Scaling of the Final Network}

The final model is large mainly because convolutional parameter counts scale approximately quadratically with channel width. If a cell topology is fixed and channel counts are multiplied by a factor $\gamma$, most pointwise and classifier-adjacent convolutional terms scale like $\gamma^2$, while batch-normalization and bias-like terms scale closer to $\gamma$. The implementation check confirms this behavior: the same architecture has about 39.24M non-auxiliary parameters at 128 initial channels and 86.94M at 192 channels. The observed ratio of parameter counts, $86.94/39.24 \approx 2.22$, is close to the quadratic prediction, $(192/128)^2 = 2.25$.
Thus, the large parameter count is not primarily caused by the 40-token architecture encoding or the low-rank search adapters; it is caused by evaluating the discovered cell with a wide 192-channel, 26-cell stand-alone classifier.

\subsection{CIFAR-10 Experiments}
\label{sec:classification-experiments}

\subsubsection{Dataset and Evaluation}

We train the selected classifier on the CIFAR-10 training split~\cite{krizhevsky2009cifar} and evaluate it on the official test split after every epoch. We report both the maximum across those evaluations and the score at the predetermined final epoch, since only the former benefits from choosing among checkpoints on that split.

\subsubsection{Search and Training Configuration}

To evaluate the selected topology independently of the search weights, we train it from scratch with 192 initial channels and a depth setting of 8, which gives 26 cells in three stages. The first $3\times3$ convolution maps the RGB input to $3C$ channels with $C=192$. At each of the two stage boundaries, a reduction cell halves the spatial resolution while the channel count doubles.

We train for 1,000 epochs using SGD with cosine learning-rate decay, a batch size of 96, and weight decay $5\times10^{-4}$. The recipe includes cutout of size 16~\cite{devries2017cutout}, AutoAugment~\cite{cubuk2019autoaugment}, label smoothing with $\epsilon=0.1$~\cite{szegedy2016rethinking}, auxiliary loss with weight 0.4, and drop-path keep probability 0.8. Stochastic weight averaging (SWA)~\cite{izmailov2018swa} begins at epoch 750. Table~\ref{tab:config} lists the search and final-training settings.

\begin{table}[H]
\centering
\caption{LRWS-NAO search and final-training configuration.}
\label{tab:config}
\begin{tabular}{@{}ll@{\qquad}ll@{}}
\toprule
\multicolumn{2}{@{}l}{Architecture search} & \multicolumn{2}{l@{}}{Stand-alone training} \\
\cmidrule(r){1-2}\cmidrule(l){3-4}
GPU & 1 NVIDIA A100 & GPU & 1 NVIDIA A100 \\
Search space & Middle, 10 operations & Depth setting / initial channels & 8 / 192 \\
Child depth / initial channels & 3 / 24 & Cells (reduction cells) & 26 (2) \\
Child epochs per iteration & 80 & Nodes per cell & 5 \\
Seed architectures & 2000 & Batch size & 96 \\
New architectures per iteration & 500 & Epochs & 1000 \\
Search iterations & 4 & Initial learning rate & 0.03 \\
Adapter rank & 16 & Weight decay & $5\times10^{-4}$ \\
Adapter learning-rate multiplier & 10.0 & Cutout size & 16 \\
Controller epochs & 1000 & Drop-path keep probability & 0.8 \\
PPO/CVaR $\alpha$ & 0.2 & Label smoothing & 0.1 \\
 & & Auxiliary loss weight & 0.4 \\
 & & SWA start & Epoch 750 \\
\bottomrule
\end{tabular}
\end{table}

\subsubsection{Selected Architecture and CIFAR-10 Accuracy}

The search selected the following encoding, with the normal cell in the first 20 positions and the reduction cell in the last 20:
\begin{quote}\small
\texttt{0 1 0 4 2 3 1 8 1 8 3 3 0 8 2 8 2 1 1 8 0 4 0 3 0 7 0 0 2 8 1 3 1 8 0 1 2 0 0 2}
\end{quote}

\begin{figure}[H]
\centering
\resizebox{\textwidth}{!}{%
\begin{tikzpicture}[
  >=Stealth,
  cellnode/.style={circle,draw,minimum size=7mm,inner sep=1pt,font=\scriptsize,fill=white},
  inputnode/.style={rectangle,rounded corners=2pt,draw,minimum width=11mm,minimum height=6mm,font=\scriptsize,fill=gray!12},
  outnode/.style={rectangle,rounded corners=2pt,draw,minimum width=14mm,minimum height=7mm,font=\scriptsize,fill=blue!7},
  edge/.style={->,line width=0.45pt},
  zeroedge/.style={->,line width=0.45pt,dashed,gray},
  edgelabel/.style={font=\tiny,fill=white,inner sep=1pt}
]
\node[font=\small\bfseries] at (4,2.25) {Normal cell};
\node[inputnode] (n0) at (0,1.1) {$s_{k-2}$};
\node[inputnode] (n1) at (0,-1.1) {$s_{k-1}$};
\node[cellnode] (n2) at (2,0.8) {$v_2$};
\node[cellnode] (n3) at (4,0.8) {$v_3$};
\node[cellnode] (n4) at (6,1.35) {$v_4$};
\node[cellnode] (n5) at (6,0) {$v_5$};
\node[cellnode] (n6) at (6,-1.35) {$v_6$};
\node[outnode] (nout) at (8,0) {concat};
\draw[edge] (n0) to[bend left=8] node[edgelabel,above] {id} (n2);
\draw[edge] (n0) to[bend right=16] node[edgelabel,below] {sep7} (n2);
\draw[edge] (n2) -- node[edgelabel,above] {sep5} (n3);
\draw[edge] (n1) to[bend right=8] node[edgelabel,below] {avg} (n3);
\draw[edge] (n1) to[bend right=42] node[edgelabel,below] {avg} (n4);
\draw[edge] (n3) -- node[edgelabel,above] {sep5} (n4);
\draw[edge] (n0) to[bend left=8] node[edgelabel,above] {avg} (n5);
\draw[edge] (n2) to[bend left=16] node[edgelabel,left] {avg} (n5);
\draw[edge] (n2) to[bend right=22] node[edgelabel,left] {id} (n6);
\draw[edge] (n1) to[bend right=8] node[edgelabel,below] {avg} (n6);
\draw[edge] (n4) -- (nout);
\draw[edge] (n5) -- (nout);
\draw[edge] (n6) -- (nout);
\node[font=\tiny] at (8,-0.65) {$[v_4,v_5,v_6]$};

\begin{scope}[xshift=10.4cm]
\node[font=\small\bfseries] at (4,2.25) {Reduction cell};
\node[inputnode] (r0) at (0,1.1) {$s_{k-2}$};
\node[inputnode] (r1) at (0,-1.1) {$s_{k-1}$};
\node[cellnode] (r2) at (2,0.8) {$v_2$};
\node[cellnode] (r3) at (4,1.35) {$v_3$};
\node[cellnode] (r4) at (4,0) {$v_4$};
\node[cellnode] (r5) at (6,-0.2) {$v_5$};
\node[cellnode] (r6) at (6,-1.45) {$v_6$};
\node[outnode] (rout) at (8,0) {concat};
\draw[edge] (r0) to[bend left=10] node[edgelabel,above] {sep7, s2} (r2);
\draw[edge] (r0) to[bend right=16] node[edgelabel,below] {sep5, s2} (r2);
\draw[edge] (r0) to[bend left=18] node[edgelabel,above] {dilsep7, s2} (r3);
\draw[zeroedge] (r0) to[bend right=4] node[edgelabel,below] {zero} (r3);
\draw[edge] (r2) -- node[edgelabel,above] {avg} (r4);
\draw[edge] (r1) to[bend right=12] node[edgelabel,below] {sep5, s2} (r4);
\draw[edge] (r1) to[bend right=8] node[edgelabel,below] {avg, s2} (r5);
\draw[edge] (r0) to[bend left=8] node[edgelabel,above] {factorized} (r5);
\draw[zeroedge] (r2) to[bend right=18] node[edgelabel,above] {zero} (r6);
\draw[edge] (r0) to[bend left=42] node[edgelabel,above] {sep3, s2} (r6);
\draw[edge] (r3) -- (rout);
\draw[edge] (r4) -- (rout);
\draw[edge] (r5) -- (rout);
\draw[edge] (r6) -- (rout);
\node[font=\tiny] at (8,-0.65) {$[v_3,v_4,v_5,v_6]$};
\end{scope}
\end{tikzpicture}}
\caption{Normal and reduction cells selected by the CIFAR-10 search. Labels identify the operation on each edge; ``s2'' indicates stride two on reduction-cell input edges. Dashed gray edges are zero operations. Each cell concatenates the intermediate nodes that are not inputs to a later node.}
\label{fig:cells}
\end{figure}
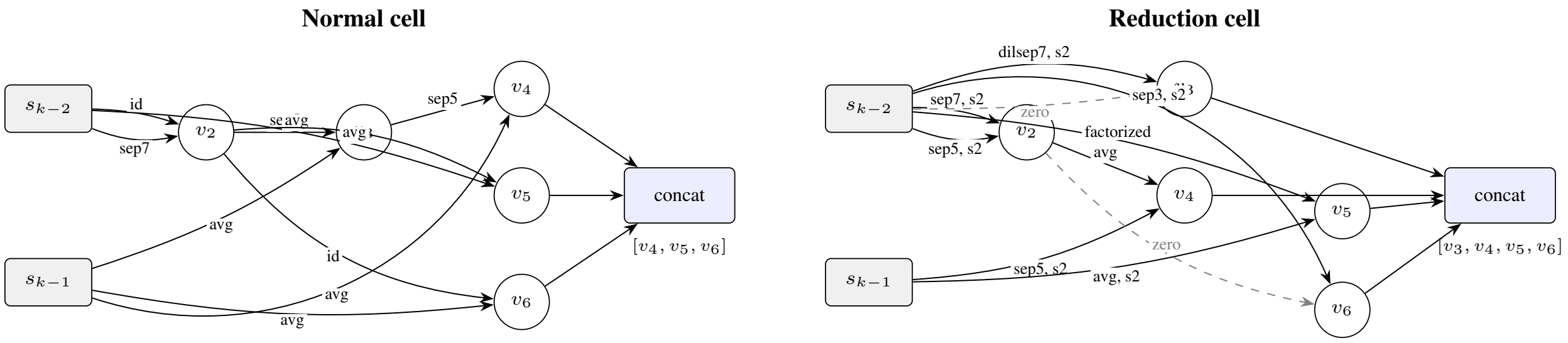

Figure~\ref{fig:cells} shows the normal and reduction cells selected for stand-alone training on CIFAR-10. Average pooling appears on five of the ten edges in the normal cell, whereas both zero operations occur in the reduction cell.

At epoch 950, the classifier reached its highest CIFAR-10 test accuracy of 98.00\%, corresponding to 2.00\% error. For comparison, accuracy was 97.78\% at the final epoch and 97.80\% after stochastic weight averaging. These values describe different checkpoints from one training run rather than variation across independent runs.

The parameter-counting function in the code computes
\begin{equation}
  \frac{1}{10^6}\sum_{\theta \in \Theta,\, \texttt{``auxiliary''}\notin name(\theta)} |\theta|,
\end{equation}
so the reported parameter value is in millions of parameters. Instantiating the final architecture gives 86.944714M non-auxiliary parameters and 87.740628M total parameters including the auxiliary head.

\begin{table}[!htbp]
\centering
\caption{LRWS-NAO CIFAR-10 result and resource summary.}
\label{tab:ours}
\begin{tabular}{lc}
\toprule
Metric & Value \\
\midrule
Best CIFAR-10 accuracy & 98.00\% \\
Best CIFAR-10 error & 2.00\% \\
Best epoch & 950 \\
Final epoch accuracy & 97.78\% \\
SWA accuracy & 97.80\% \\
Parameters excluding auxiliary head & 86.944714M \\
Parameters including auxiliary head & 87.740628M \\
Search time & 35.42 hours, 1.48 A100 GPU-days \\
Training time to best checkpoint & 133.40 hours, 5.56 A100 GPU-days \\
Search + training time to best checkpoint & 7.03 A100 GPU-days \\
\bottomrule
\end{tabular}
\end{table}

\subsubsection{Empirical Training and Search Curves}

\begin{figure}[ht]
\centering
\begin{minipage}{0.49\textwidth}
  \centering
  \includegraphics[width=\linewidth]{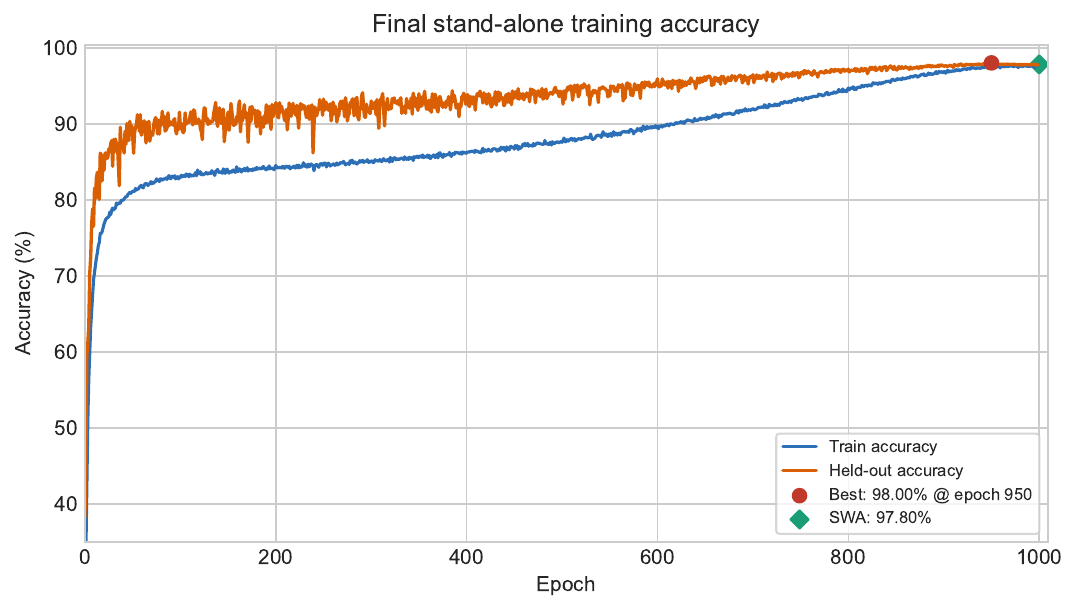}
\end{minipage}
\hfill
\begin{minipage}{0.49\textwidth}
  \centering
  \includegraphics[width=\linewidth]{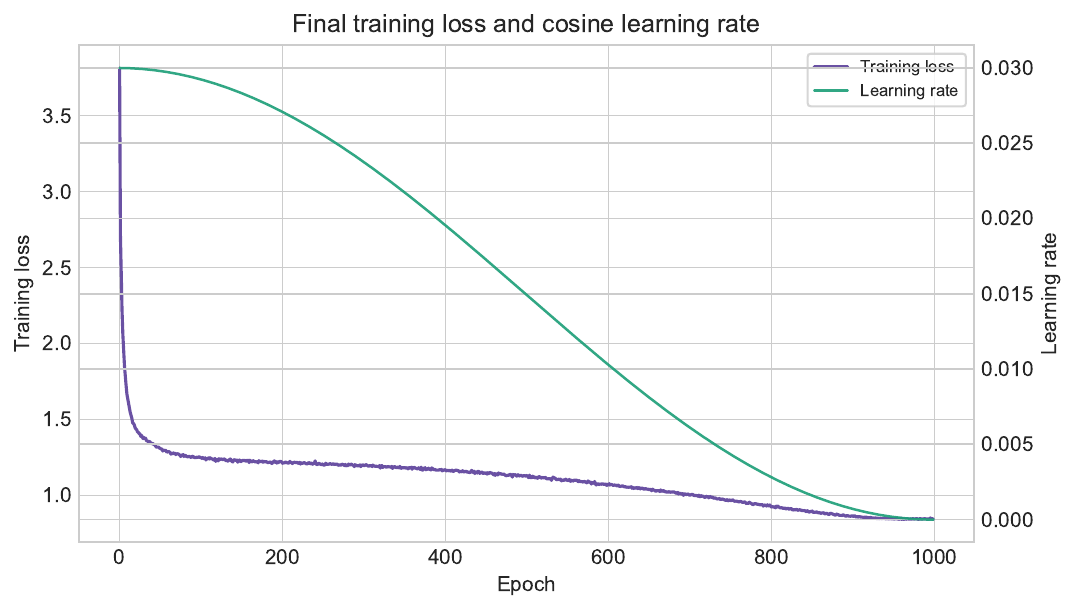}
\end{minipage}
\caption{Final training curves. Left: train and held-out accuracy, with the best non-SWA checkpoint and SWA evaluation marked. Right: epoch-level training loss and the cosine learning-rate schedule.}
\label{fig:final_curves}
\end{figure}

Figure~\ref{fig:final_curves} places the peak at epoch 950 in the context of the full training trajectory. The SWA result is marked separately because it evaluates an average of late-stage weights rather than the weights from any single epoch.

\begin{figure}[ht]
\centering
\begin{minipage}{0.49\textwidth}
  \centering
  \includegraphics[width=\linewidth]{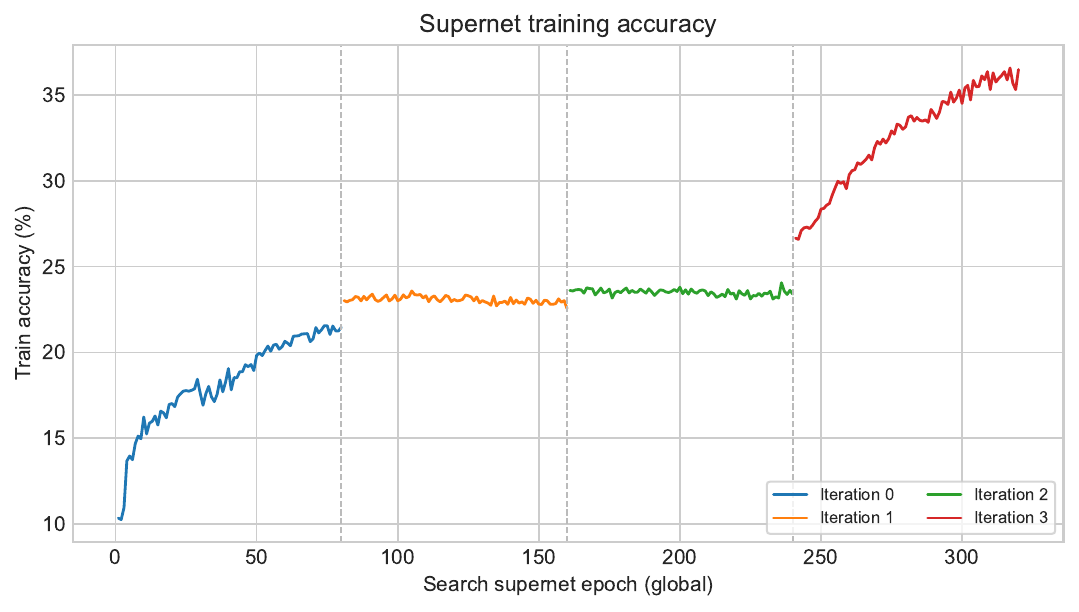}
\end{minipage}
\hfill
\begin{minipage}{0.49\textwidth}
  \centering
  \includegraphics[width=\linewidth]{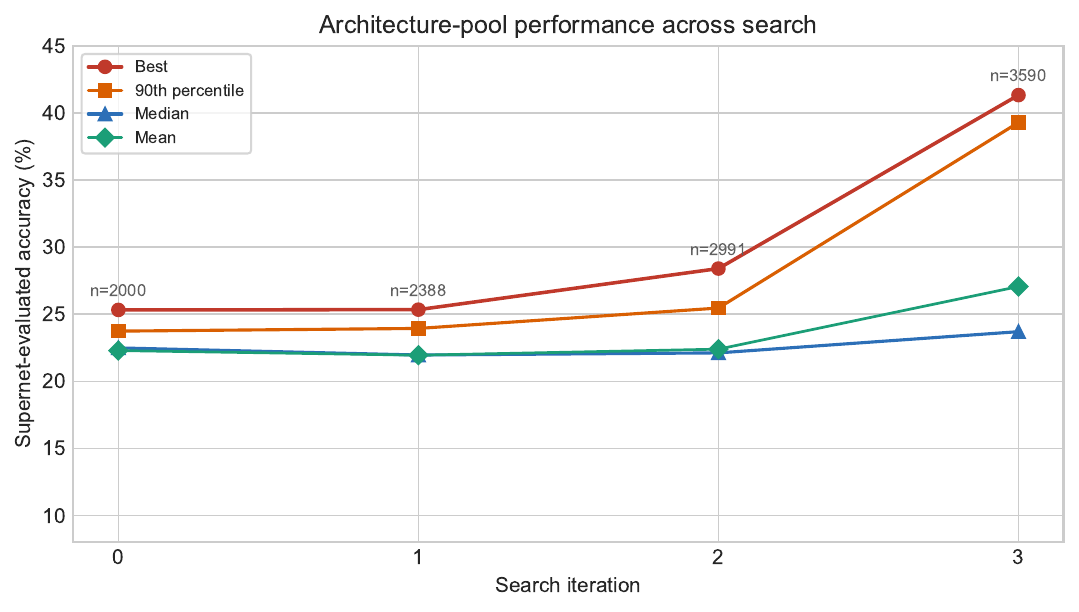}
\end{minipage}
\caption{Search-stage curves. Left: search-supernet training accuracy by iteration. Right: mean, median, 90th percentile, and best architecture-pool scores under the trained search supernet.}
\label{fig:search_curves}
\end{figure}

Figure~\ref{fig:search_curves} summarizes the search run. The left plot shows training accuracy under low-rank weight sharing across the four 80-epoch search iterations. The right plot shows that the best and upper-tail supernet-evaluated architecture scores rise sharply in the final search iteration.

\begin{figure}[ht]
\centering
\begin{minipage}{0.49\textwidth}
  \centering
  \includegraphics[width=\linewidth]{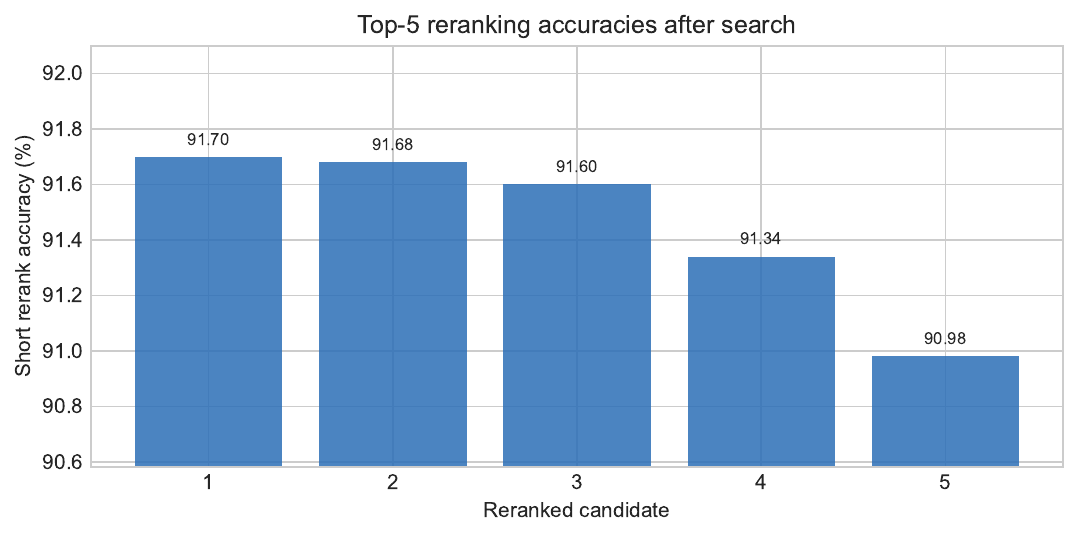}
\end{minipage}
\hfill
\begin{minipage}{0.49\textwidth}
  \centering
  \includegraphics[width=\linewidth]{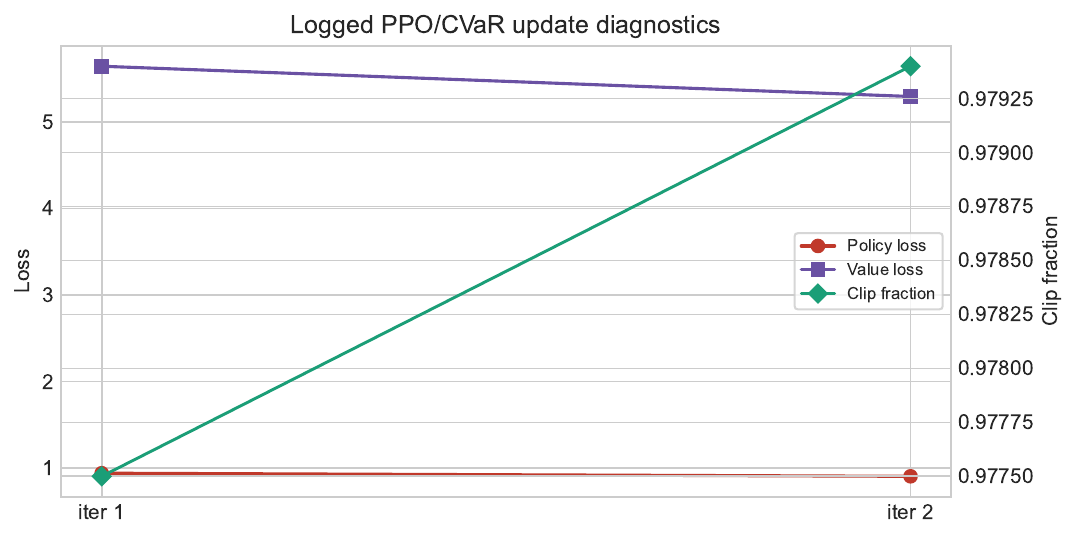}
\end{minipage}
\caption{Final search diagnostics. Left: top-5 short reranking accuracies after search. Right: PPO/CVaR losses and clip fractions for the policy updates.}
\label{fig:rerank_ppo}
\end{figure}

Figure~\ref{fig:rerank_ppo} shows the last-stage reranking and the PPO/CVaR diagnostics. The PPO statistics are reported for the two policy-update phases in iterations 1 and 2. The high clip fractions indicate that the sampled latent policy update was aggressive, which is why the implementation's early-stop rule halted the additional PPO update attempts.

\subsubsection{Search and Training Dynamics}

Table~\ref{tab:training_dynamics} shows how the interpretation changes with training duration: the classifier first reaches 90\% accuracy at epoch 44, but moving from 95\% to 98\% takes hundreds more epochs. Over the final 50 epochs, test accuracy averages 97.80\% with a standard deviation of 0.05 percentage points; over the final 100, the corresponding values are 97.77\% and 0.09 points. The peak is therefore close to the late-run average, although variation between neighboring checkpoints does not estimate variation across independent runs.

\begin{table}[ht]
\centering
\caption{Training milestones for the final CIFAR-10 run.}
\label{tab:training_dynamics}
\begin{tabular}{lcc}
\toprule
Event & Epoch & Held-out accuracy (\%) \\
\midrule
First accuracy $\ge 90\%$ & 44 & 90.01 \\
First accuracy $\ge 95\%$ & 513 & 95.07 \\
First accuracy $\ge 97\%$ & 782 & 97.15 \\
First accuracy $\ge 97.5\%$ & 839 & 97.51 \\
First accuracy $\ge 97.8\%$ & 918 & 97.86 \\
Best non-SWA checkpoint & 950 & 98.00 \\
Final epoch & 999 & 97.78 \\
SWA evaluation & -- & 97.80 \\
\bottomrule
\end{tabular}
\end{table}

The largest change in proxy scores occurs in the final search iteration (Table~\ref{tab:search_dynamics}). Across the first three pools, the best score ranges from 25.32\% to 28.40\%, whereas in the fourth pool the mean reaches 27.06\% and the best score reaches 41.32\%. Because each iteration updates the shared weights and reevaluates the accumulated pool, the increase cannot be attributed to better candidate architectures alone: the evaluator also changes.

\begin{table}[ht]
\centering
\caption{Search dynamics across architecture-pool refinement. Scores are supernet proxy accuracies.}
\label{tab:search_dynamics}
\begin{tabular}{cccccc}
\toprule
Iteration & Pool size & Supernet train max (\%) & Pool mean (\%) & Pool best (\%) & Next candidates \\
\midrule
0 & 2000 & 21.56 & 22.29 & 25.32 & 388 \\
1 & 2388 & 23.58 & 21.95 & 25.34 & 603 \\
2 & 2991 & 24.06 & 22.40 & 28.40 & 599 \\
3 & 3590 & 36.56 & 27.06 & 41.32 & final rerank \\
\bottomrule
\end{tabular}
\end{table}

Validation pairwise accuracy measures how often the predictor orders two architectures in the same way as the supernet. It reaches roughly 0.73, 0.85, and 0.91 in iterations 0, 1, and 2, respectively, although in the latter two iterations it declines after an earlier peak. This makes validation ranking a possible stopping criterion for predictor training. The decoder's near-zero reconstruction Hamming distances answer a different question: they show that it can recover an input sequence, not how far it moves when proposing a new one.

In each of the two hybrid rounds, approximately 98\% of sampled likelihood ratios fell outside the PPO clipping interval in the update diagnostics, triggering the stopping rule before a second call to the policy optimizer routine. The number of candidates in the new list grew from 274 to 362 after PPO decoding in the first round and from 305 to 376 in the second. Thus PPO added 88 and 71 entries to the respective lists before fallback generation and final filtering; these counts do not establish how many of those entries would outperform NAO proposals under independent training.

The final classifier takes about 8.41 minutes per training epoch at batch size 96, with roughly 23 additional seconds for evaluation. The minibatch timing corresponds to about 104 training images per second and should not be interpreted as inference throughput. During search, scoring one candidate with shared weights takes about 2.79 minutes on average, compared with about 24 minutes to train and evaluate one shortlisted proxy for 50 epochs.

\subsubsection{Operation Usage Motifs}

Average pooling is frequent not only in the selected cells (Figure~\ref{fig:cells}) but also among the other highly ranked candidates. Table~\ref{tab:op_motifs} reports edge-operation frequencies for the selected architecture, the final top-5 reranked architectures, and the top 100 architectures in the last supernet pool. The selected architecture uses average pooling on 35\% of its edges, identity on 15\%, and separable convolutions on 35\%. The top-5 candidates show a similar pattern, with average pooling as the dominant operation. This suggests that the search favors smoothing and skip-style pathways combined with a smaller number of medium-width separable convolutions.

\begin{table}[ht]
\centering
\caption{Operation frequencies in selected and high-ranked architectures. Percentages are over all encoded edges in normal and reduction cells.}
\label{tab:op_motifs}
\begin{tabular}{lccc}
\toprule
Operation & Selected arch (\%) & Final top-5 (\%) & Final pool top-100 (\%) \\
\midrule
Average pooling & 35.0 & 36.0 & 32.5 \\
Identity / factorized reduction & 15.0 & 16.0 & 22.4 \\
SepConv3 & 5.0 & 5.0 & 18.3 \\
SepConv5 & 20.0 & 14.0 & 8.8 \\
SepConv7 & 10.0 & 8.0 & 3.7 \\
Zero & 10.0 & 11.0 & 10.2 \\
DilSepConv3 & 0.0 & 4.0 & 1.8 \\
DilSepConv7 & 5.0 & 6.0 & 1.7 \\
\bottomrule
\end{tabular}
\end{table}

\subsubsection{Supernet Ranking Versus Short-Train Reranking}

To assess whether shared weights preserve the ordering that matters for selection, we compare the five entries passed to independent reranking with their validation accuracies after 50 training epochs. Each row of Table~\ref{tab:ranking_correlation} pairs the independent result with the score used to shortlist that architecture, rather than with a later occurrence of the same encoding in the archive. Absolute scores are on different scales because the training protocols differ; it is their ordering that matters here.

\begin{table}[H]
\centering
\caption{Shared-weight and independent validation scores for the five CIFAR-10 finalists. Rank 1 denotes the highest accuracy within the shortlist. The two identical supernet scores occupy positions 2 and 3 and receive their average rank, 2.5.}
\label{tab:ranking_correlation}
\begin{tabular}{cccc}
\toprule
Short-train rank & Supernet rank & Supernet score (\%) & 50-epoch accuracy (\%) \\
\midrule
1 & 2.5 & 41.28 & 91.70 \\
2 & 4 & 41.26 & 91.68 \\
3 & 1 & 41.32 & 91.60 \\
4 & 2.5 & 41.28 & 91.34 \\
5 & 5 & 41.22 & 90.98 \\
\bottomrule
\end{tabular}
\end{table}

For the five finalists, assigning average ranks to tied scores gives a Spearman rank correlation of $\rho_{\mathrm{Spearman}} \approx 0.359$, whereas comparing the order of candidate pairs and accounting for ties gives a Kendall coefficient of $\tau_b \approx 0.316$.
Of the ten candidate pairs, six retain their order, three reverse it, and one is tied under the supernet. The proxy leader finishes third after independent training, while the two candidates tied at 41.28\% separate to 91.70\% and 91.34\%. This example explains why a short independent run can change the final choice even when shared weights are useful for narrowing the pool. Because the five finalists were selected for high proxy scores, neither coefficient measures ranking reliability across the full search space.

\subsubsection{Comparison with NAS Literature}

Table~\ref{tab:comparison} places the CIFAR-10 result alongside representative NAS methods. The closest reported error is that of large NAONet: 1.93\% with 144.6M parameters, compared with our peak of 2.00\% with 86.9M. Several alternatives are considerably smaller; for example, ProxylessNAS reports 2.08\% with 5.7M parameters. Since the training recipes, checkpoint selection rules, and hardware are not matched across papers, these figures describe different accuracy, size, and search-cost trade-offs rather than an isolated effect of the search algorithm.

\begin{table}[!htbp]
\centering
\caption{CIFAR-10 comparison with representative NAS methods. Lower error is better. ``Ops'' denotes the number of candidate operations per searched edge for cell-based methods; for ProxylessNAS it denotes block-level mobile choices, so it is not directly identical to a cell-edge operation count. ``Space'' is an approximate discrete search-space size under the method's reported cell/block convention. Search cost follows the convention of reporting architecture-search cost, not final stand-alone training cost.}
\label{tab:comparison}
\resizebox{\textwidth}{!}{%
\begin{tabular}{lcccccc}
\toprule
Method & Error (\%) & Params (M) & Ops & Space & Search cost & Search type \\
\midrule
NAONet + cutout, large \cite{luo2018nao} & 1.93 & 144.6 & 5 & $\sim2.6{\times}10^{25}$ & 200 GPU-days & NAO \\
\textbf{LRWS-NAO (ours)} & \textbf{2.00} & \textbf{86.9} & \textbf{10} & $\mathbf{2.7{\times}10^{31}}$ & \textbf{1.48 A100 GPU-days} & NAO + one-shot \\
ProxylessNAS + cutout \cite{cai2019proxylessnas} & 2.08 & 5.7 & 7 & block-level & 4 GPU-days & gradient-based \\
AmoebaNet-B + cutout \cite{real2019regularized} & 2.13 & 34.9 & 5 & $\sim2.6{\times}10^{25}$ & 3150 GPU-days & evolution \\
P-DARTS, large \cite{chen2019pdarts} & 2.25 & 10.5 & 8 & $\sim1.1{\times}10^{18}$ & 0.3 GPU-days & gradient-based \\
NAONet + cutout, compact \cite{luo2018nao} & 2.48 & 10.6 & 5 & $\sim2.6{\times}10^{25}$ & 200 GPU-days & NAO \\
P-DARTS + cutout \cite{chen2019pdarts} & 2.50 & 3.4 & 8 & $\sim1.1{\times}10^{18}$ & 0.3 GPU-days & gradient-based \\
PC-DARTS + cutout \cite{xu2020pcdarts} & 2.57 & 3.6 & 8 & $\sim1.1{\times}10^{18}$ & 0.1 GPU-days & gradient-based \\
Balanced NAO \cite{luo2019balanced} & 2.60 & 3.8 & 5 & $\sim2.6{\times}10^{25}$ & -- & one-shot NAO \\
DARTS, second order + cutout \cite{liu2019darts} & 2.76 & 3.3 & 8 & $\sim1.1{\times}10^{18}$ & 4 GPU-days & gradient-based \\
NAONet-WS + cutout \cite{luo2018nao} & 2.93 & 2.5 & 5 & $\sim2.6{\times}10^{25}$ & 0.3 GPU-days & one-shot NAO \\
\bottomrule
\end{tabular}
}
\end{table}

For LRWS-NAO, Table~\ref{tab:cardinality} gives the exact number of encodings. For the NAO and AmoebaNet families, the comparison uses an analogous estimate with two cells, five intermediate nodes per cell, and five operations. The DARTS estimate uses four nodes and eight operations during search, retaining two nonzero incoming edges per node in the discrete cell. Because the methods count equivalent cells differently, these estimates indicate the scale of search rather than a common number of distinct networks.

\section{Segmentation}
\label{sec:seg}

Semantic segmentation requires a model to use the surrounding scene to interpret local features without losing track of where those features occur. We keep the overall structure of the model fixed and search for different ways to combine local detail with broader context. As in classification, we use low-rank weight sharing: candidates train common components together, while selected operations can adapt their shared weights to different input sources.

\subsection{Architecture Search}
\subsubsection{Cell Space and Shared Weights}
Unlike the classification search, which specifies separate normal and reduction cells, the segmentation search repeats a single cell while following the fixed 12-layer Auto-DeepLab resolution path $(0,0,0,1,2,1,2,2,3,3,2,1)$. Level $\ell$ denotes output stride $4\cdot 2^\ell$: for example, level zero retains a feature map at one quarter of the input height and width, while the final level one retains one eighth. Within each cell, five successive nodes each transform two distinct predecessors and add their outputs; concatenating the five sums gives the cell output. Figure~\ref{fig:seg-arch}A shows the fixed path and decoder. Changing an edge alters which feature is processed and how, without moving the stage at which the encoder downsamples.

We consider two ways to choose a cell on this path. In the \emph{global} search, PPO selects the source and operation for each of ten edges, masking choices that would violate the cell constraints. It can choose among seven operations: $3{\times}3$ and $5{\times}5$ separable convolution, $3{\times}3$ and $5{\times}5$ atrous convolution, identity, and $3{\times}3$ average or maximum pooling. At most two edges may use identity, and selection imposes a 24M-parameter limit at filter multiplier $F=32$. In the \emph{reference-guided} search, by contrast, a single categorical action chooses among 128 cells: the reference and variants with one or two edge replacements. Each replacement chooses a valid source and one of the four convolutions. Swapping the two branches of a sum does not create another cell in this library, and the parameter cap is the reference size of 10,147,427 at $F=20$. This narrower choice lets us test small changes without increasing the model budget.

As in classification, the shared convolution can adjust to the source of its input. Each minibatch activates one topology along the entire backbone; a shared pointwise or atrous kernel $W$ applied to source $s$ becomes $W+B_sA_s$, with the product reshaped to the kernel dimensions. The base kernel remains trainable, while each source starts from zero correction and learns an adapter of rank at most four. Source-specific batch normalization (BN) accounts for differences in feature statistics, and the adapters use twice the base learning rate with no weight decay. For independent evaluation, we retain the selected topology and train ordinary convolutional weights from scratch.

\subsubsection{Image-Level Risk and Selection}
A segmentation cell can improve performance on a class without improving the most difficult images. To distinguish these cases, let $M(a)$ denote the mean IoU across evaluation classes for cell $a$, obtained by accumulating pixel counts over the evaluation panel before computing each class's intersection over union. For each image $i$, let $q_i(a)$ instead be its mean IoU over classes present in its ground truth, excluding void pixels. For $n$ images, sort these image scores from lowest to highest and set $h=0.2n$ and $k=\lfloor h\rfloor$. The segmentation controller receives the reward
\begin{equation}
 T_{0.2}(a)=\frac{\sum_{i=1}^{k}q_{(i)}(a)+(h-k)q_{(k+1)}(a)}{h},
 \qquad R(a)=0.8M(a)+0.2T_{0.2}(a).
 \label{eq:seg-reward}
\end{equation}
Consider a rollout panel of 64 images. Its lowest fifth has mass 12.8, so the tail term includes the 12 lowest image scores in full and gives the thirteenth a weight of 0.8 before dividing their sum by 12.8. Note that the tail is computed \emph{within each cell} before PPO compares cells: the discrete policy uses the whole rollout for its update, with each reward measured relative to a critic baseline conditioned on training progress, relative learning rate, and recent loss. In classification, by contrast, the actor selects a tail \emph{across proposed moves} after forming its PPO surrogate; it does not rank images.

Algorithm~\ref{alg:seg} specifies the reference-guided search. Weight updates, controller rewards, and final cell selection use disjoint training partitions so that a candidate is not selected on the same images used to fit the policy.

\begin{pagedalgorithm}{Reference-guided segmentation cell search and independent retraining}
\label{alg:seg}
\begin{algorithmic}[1]
\Require Reference cell $a_0$, fixed resolution path, and the three training partitions
\State Construct the 128-cell library under the $F=20$ parameter cap
\State Initialize the $F=8$ supernet, rank-four adapters, categorical actor, and critic
\For{epoch $e=1,\ldots,100$}
  \For{each minibatch of six $321{\times}321$ weight-training crops}
    \State Use $a_0$ every tenth minibatch. Otherwise sample uniformly for $e\leq 50$,
    \Statex \hspace{\algorithmicindent}\hspace{\algorithmicindent}then from an equal mixture of uniform sampling and the current policy
    \State Update the active path by pixelwise cross-entropy
  \EndFor
  \If{$e>50$}
    \State Sample 16 cells from the policy and draw a common controller panel
    \State Calibrate BN on eight training minibatches per cell and measure $R(a)$
    \State Standardize $R(a)-V(s)$ across the rollout and update actor and critic
    \State Use up to four PPO epochs with ratio clipping at $1\pm0.2$ and a KL limit of $0.02$
  \EndIf
\EndFor
\State Score all 128 cells on the selection split at the final supernet weights
\State Retain $a_0$, the mIoU leader, a reward leader among the remaining cells,
\Statex \hspace{\algorithmicindent}and two high-scoring challengers selected for edge diversity
\State Train the five cells independently at $F=20$ on the combined weight-training and controller partitions
\State At 50k updates, retain $a_0$ and the two highest-mIoU challengers
\State At 150k, select the best challenger clearing a 0.5-point margin over $a_0$
\Statex \hspace{\algorithmicindent}at the probe seed, falling back to $a_0$ otherwise
\State Retrain the selected cell and $a_0$ from scratch on the full training set
\end{algorithmic}
\end{pagedalgorithm}

\subsection{Cityscapes Experiments}
\label{sec:seg-experiments}

\subsubsection{Dataset and Evaluation Protocol}

We divide the 2,975 finely annotated Cityscapes training images~\cite{cordts2016cityscapes} into disjoint sets for weight optimization (1,785), controller rewards (595), and architecture selection (595). Within a rollout, all 16 sampled cells see the same 64 reward images at the same frozen supernet weights. Otherwise, one cell could appear stronger simply because its panel contained easier scenes. Before scoring each cell, we also recalibrate BN on common weight-training crops and restore the previous statistics afterward, so an earlier cell cannot influence a later score through those statistics. Final proxy ranking uses all 595 selection images and 32 calibration minibatches; the separate 500-image official validation split is used only after full retraining, including for checkpoint selection. We report mIoU across the 19 evaluation classes and compute the image-level tail as the mean of the 100 lowest image scores on this split.

Independent probes use the 2,380 training images outside the selection partition. For the final comparison, we trained the Auto-DeepLab-S reference architecture \cite{liu2019autodeeplab} and the selected challenger from scratch in the same implementation. For both models, we fixed $F=20$, seed 0, 500,000 optimizer updates, batch size 16, and $769{\times}769$ training crops. Both networks used SGD with momentum 0.9, weight decay $10^{-4}$, a 1,000-update warmup to a learning rate of 0.05, polynomial decay with exponent 0.9, BN momentum 0.01, and zero drop path. We evaluated every 5k updates and selected each checkpoint by single-scale validation mIoU.

\subsubsection{Selected Cells and Independent Training}

\begin{figure}[!htbp]
\centering
\includegraphics[width=\linewidth]{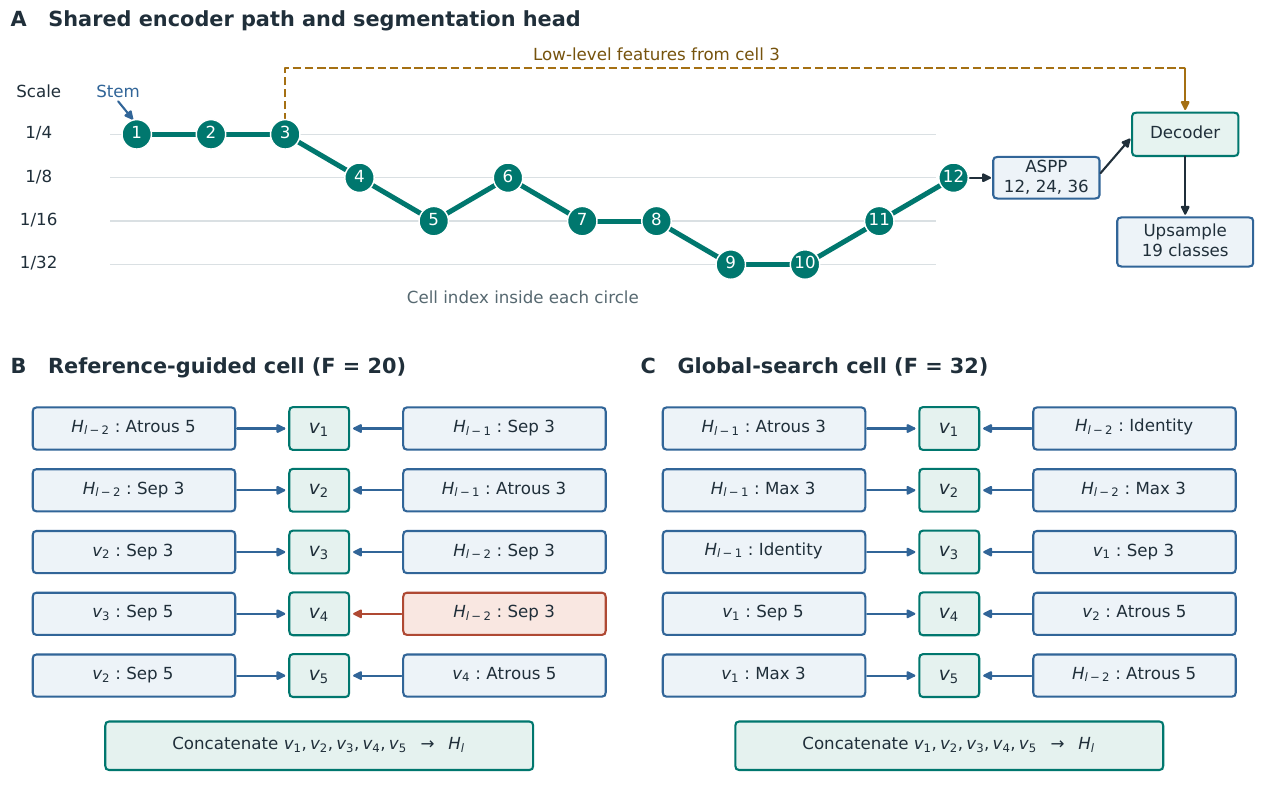}
\caption{Selected segmentation architectures. (A) Both encoders follow the indicated 12-cell path. Cell 12 feeds atrous spatial pyramid pooling (ASPP) at stride eight, while cell 3 supplies stride-four features to the decoder. Predictions are upsampled to the input resolution. (B,C) Each row depicts the two incoming edges of one sum $v_i$. An edge specifies its source (the preceding cell $H_{l-1}$, the older cell $H_{l-2}$, or an earlier $v_j$) and its operation. Sep, Atrous, and Max denote separable convolution, atrous convolution, and max pooling; the number denotes square kernel width. The highlighted branch in (B) replaces $\mathrm{Sep}_{5}(v_1)$ in the reference with $\mathrm{Sep}_{3}(H_{l-2})$. This topology is reused at every resolution.}
\label{fig:seg-arch}
\end{figure}

Figure~\ref{fig:seg-arch} shows the cell selected by the reference-guided search (B) and the cell selected by the global search (C). Only the reference-guided cell was compared with our reproduced Auto-DeepLab-S under matched training conditions.

In the global search, 758 distinct candidates were evaluated on the selection split. Among cells matching or exceeding the reference proxy mIoU, the cell with the highest risk-adjusted reward attained 34.46\% proxy mIoU, compared with 33.83\% for the reference. After 740,000 training updates at $F=32$, the selected 19.48M-parameter network reached \textbf{75.079\% single-scale validation mIoU}. This configuration uses scheduled drop path up to 0.1. A reference run at the same width and update budget would be needed to estimate the gain attributable to this cell.

\begin{figure}[!htbp]
\centering
\includegraphics[width=\linewidth]{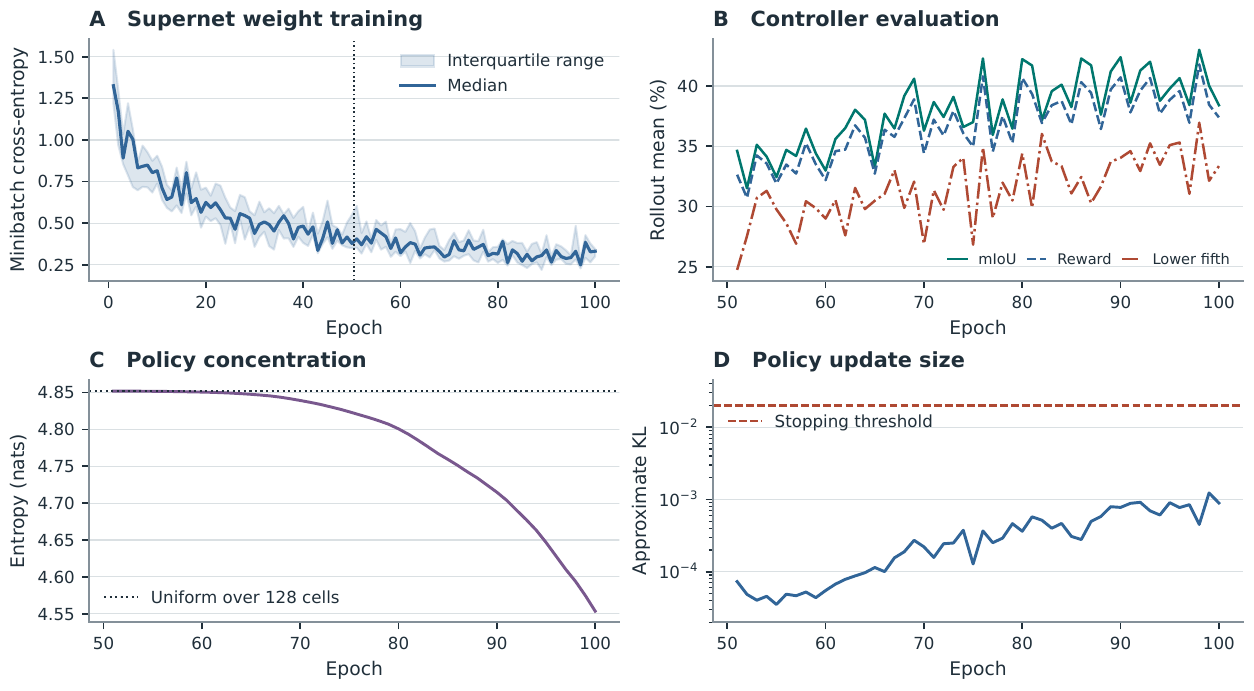}
\caption{Reference-guided weight and policy optimization. (A) Median and interquartile range of sampled minibatch cross-entropies in each epoch; the vertical line marks the end of warmup. (B) Each point averages 16 sampled cells on a common panel of 64 images. Panels vary by rollout, so changes between epochs reflect sampling variation as well as optimization. (C,D) Categorical policy entropy and approximate KL divergence. The dotted entropy baseline is $\log 128$, the entropy of uniform sampling over the library.}
\label{fig:seg-search}
\end{figure}

The reference-guided search (Algorithm~\ref{alg:seg}) used 29,700 supernet weight updates and 50 PPO rollouts. The 100-epoch supernet and controller stage took 4.01 hours on one A100 (0.167 GPU-days), excluding final scoring of the 128 cells, 550k independent probe updates, and final retraining. The probe budget comprises five cells trained to 50k updates and three continued for another 100k each. In Figure~\ref{fig:seg-search}, all 50 rollouts completed four PPO passes, and the largest reported approximate KL divergence remained below 0.00124, compared with the stopping threshold of 0.02. This indicates that the policy distribution changed little in each update; it does not tell us whether the resulting cell rankings will survive independent training.

\begin{figure}[!htbp]
\centering
\includegraphics[width=\linewidth]{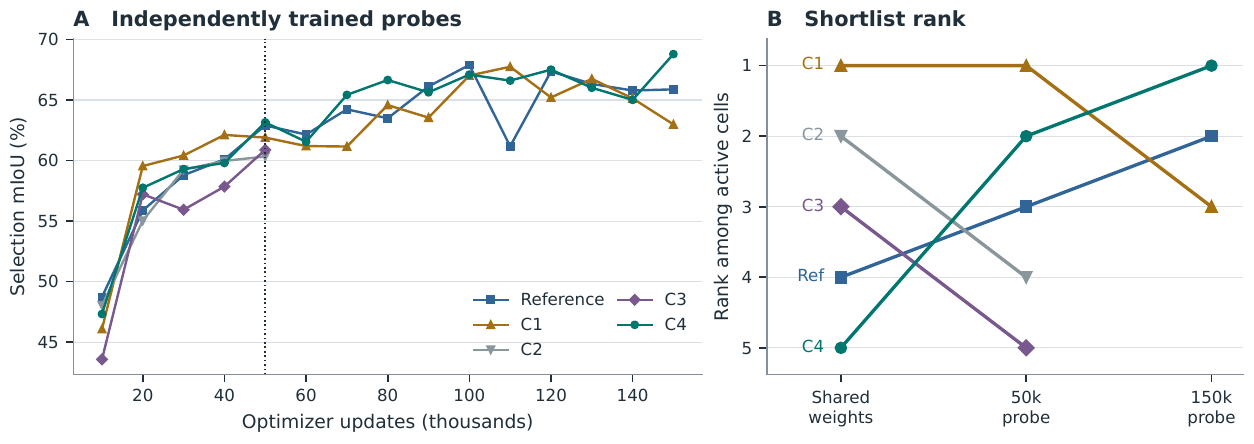}
\caption{Selection on held-out training images. (A) Each marker is one evaluation of an independently trained cell on the same 595 images. Two cells stop at 50k updates. (B) Ranks among the five shortlisted cells, then among the three survivors. Probe ranks use the last three evaluations at each budget, not the last marker in (A). The reference is our Auto-DeepLab-S cell and advances by design.}
\label{fig:seg-probes}
\end{figure}

The independent probes illustrate why training duration matters. Candidate 1 led at 50k updates, yet by 150k its score was 1.04 points below the reference, whereas candidate 4 had moved from fifth under shared weights to second at 50k and first among the three remaining cells at 150k (Table~\ref{tab:seg-probes} and Figure~\ref{fig:seg-probes}). Its 0.61-point advantage over the reference exceeded the 0.5-point promotion margin on the available seed. The difference in topology is confined to one branch of the fourth node: a $3{\times}3$ separable convolution of the older incoming cell state replaces a $5{\times}5$ separable convolution of the first intermediate node (Figure~\ref{fig:seg-arch}B). Candidate 4 advanced without the planned second-seed confirmation.

\begin{table}[!htbp]
\centering
\caption{Reference-guided shortlist on the 595-image selection partition of the training set. Probe mIoU (\%) is the mean of the last three evaluations, spaced 10k updates apart. A dash denotes a cell eliminated after 50k.}
\label{tab:seg-probes}
\begin{tabular}{lrrr}
\toprule
Cell & Shared-weight mIoU (\%) & 50k probe (\%) & 150k probe (\%) \\
\midrule
Auto-DeepLab-S (cell 0) & 44.52 & 60.60 & 66.00 \\
1 & 44.76 & 61.48 & 64.96 \\
2 & 44.70 & 59.85 & -- \\
3 & 44.65 & 58.23 & -- \\
4 & 44.48 & 60.75 & \textbf{66.61} \\
\bottomrule
\end{tabular}
\end{table}

\subsubsection{Matched Comparison with Reproduced Auto-DeepLab-S}
The reproduced Auto-DeepLab-S contains 10,147,427 parameters. Candidate 4 contains 10,121,187, a reduction of 26,240.

\begin{table}[!htbp]
\centering
\caption{Cityscapes validation performance on all 500 images under matched $F=20$ training. Auto-DeepLab-S denotes our reproduction throughout. Its best checkpoint occurs at 475k updates; candidate 4 peaks at 490k. The tail is the mean IoU of the lowest-scoring 100 images for each model. Differences are in percentage points.}
\label{tab:seg-matched}
\small
\begin{tabular}{llrrr}
\toprule
Inference & Checkpoint and metric & Auto-DeepLab-S (\%) & Candidate 4 (\%) & Difference \\
\midrule
Single scale & Best, mIoU & 73.942 & 74.285 & $+0.343$ \\
Single scale & Final 500k, mIoU & 73.117 & 74.132 & $+1.016$ \\
Single scale & Best, image tail & 55.365 & 54.998 & $-0.368$ \\
Single scale & Final 500k, image tail & 55.869 & 54.879 & $-0.989$ \\
\midrule
Five scales and flips & Best, mIoU & 75.989 & 76.323 & $+0.334$ \\
Five scales and flips & Best, image tail & 56.411 & 56.037 & $-0.374$ \\
\bottomrule
\end{tabular}
\end{table}

\begin{figure}[!htbp]
\centering
\includegraphics[width=\linewidth]{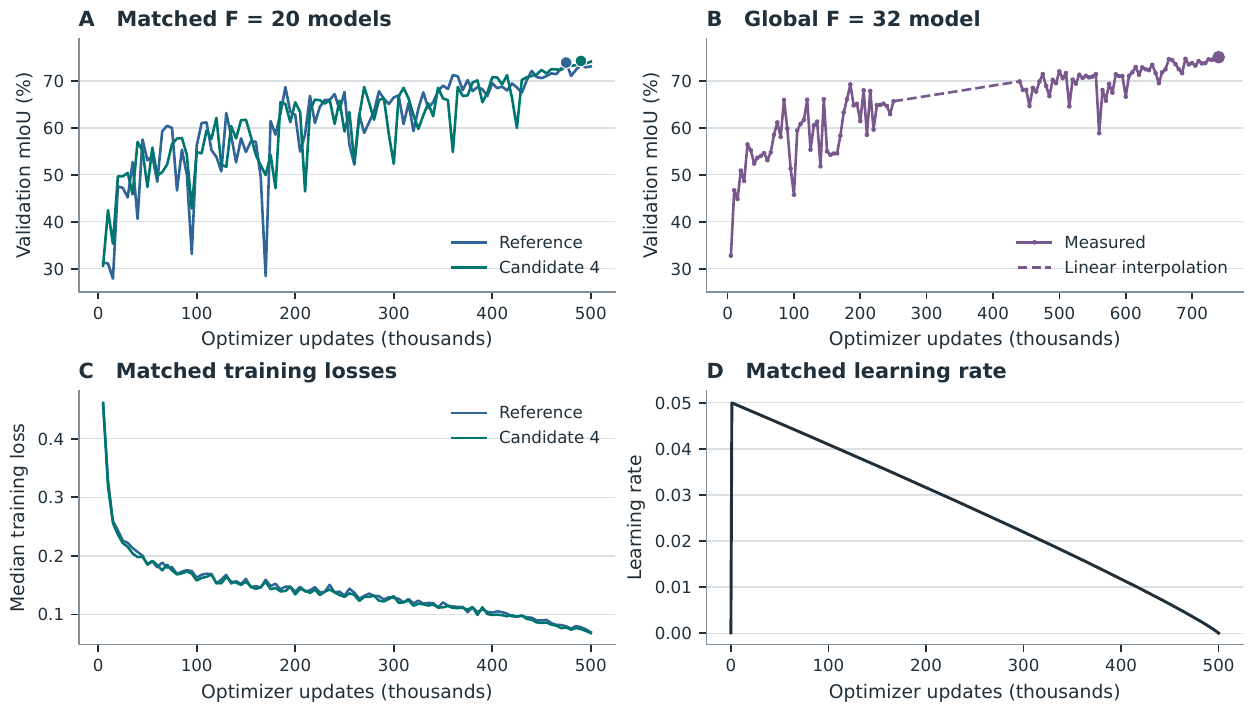}
\caption{Training and validation from scratch. (A) All 100 single-scale evaluations of candidate 4 and the reproduced Auto-DeepLab-S (Reference) on the 500 Cityscapes validation images at $F=20$. Dots mark the best checkpoints. (B) The $F=32$ global model used a different schedule. The dashed segment linearly interpolates between measured validation mIoU at 250k and 440k updates; only measured checkpoints contribute to reported results. (C) Median of sampled training losses in nonoverlapping 5k-update intervals. (D) The two $F=20$ models share the same learning-rate schedule.}
\label{fig:seg-retraining}
\end{figure}

With single-scale mixed-precision inference, candidate 4 exceeds our Auto-DeepLab-S baseline by 0.343 mIoU points at the best checkpoints and by 1.016 points at the common final update (Figure~\ref{fig:seg-retraining}A). For the augmented evaluation, we average class probabilities at scales $\{0.5,0.75,1,1.25,1.5\}$ with and without horizontal reflection, giving ten forward passes per image. Under this shared protocol the advantage is 0.334 points. A full-precision single-scale check gives 73.943\% for the reference and 74.288\% for candidate 4 using stored BN statistics, or 73.909\% and 74.181\% after recalibration on training images.

\begin{figure}[!htbp]
\centering
\includegraphics[width=\linewidth]{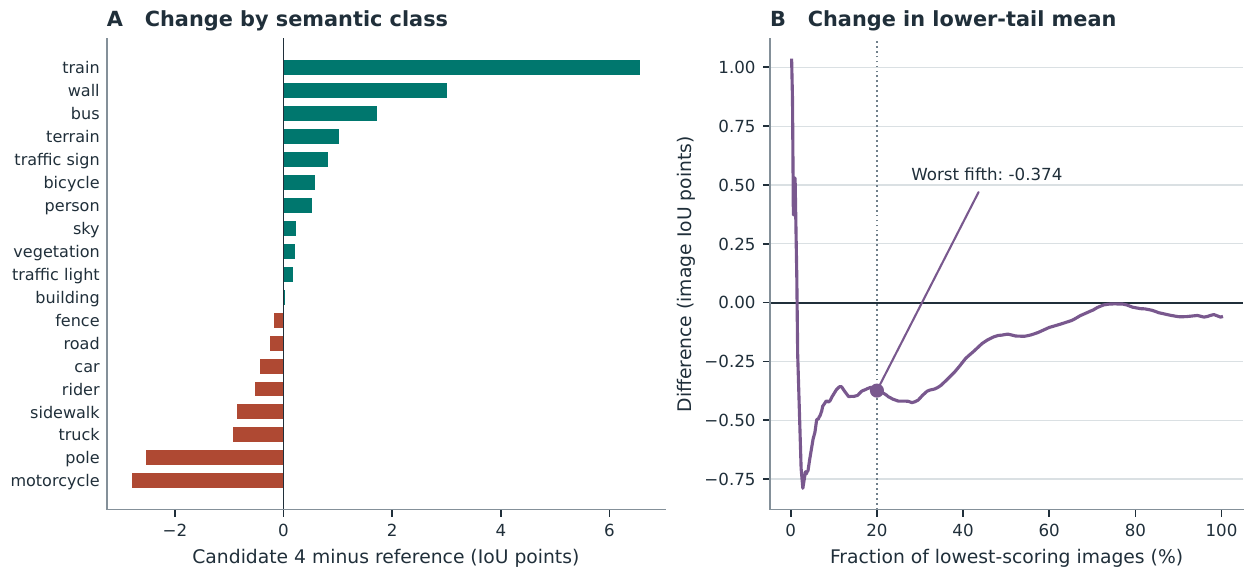}
\caption{Where the matched $F=20$ models differ under five-scale and flip inference at their best single-scale checkpoints. (A) Per-class IoU differences computed from the two aggregate confusion matrices. Their unweighted mean is $+0.334$ points. (B) For each fraction of the 500 validation images, we rank images independently within each model, average the lowest scores, and subtract the average for the reproduced Auto-DeepLab-S. The highlighted point is the prespecified 20\% tail. The two tails need not contain the same images.}
\label{fig:seg-class}
\end{figure}

The aggregate and image-tail metrics move in opposite directions under both inference protocols (Table~\ref{tab:seg-matched}). With multi-scale and flip inference, candidate 4 improves 11 of the 19 class IoUs. The largest gains are for the train and wall classes ($+6.55$ and $+3.01$ points), while motorcycle and pole show the largest losses ($-2.78$ and $-2.53$). Yet the average over each model's 100 lowest-scoring images falls from 56.411\% to 56.037\% (Figure~\ref{fig:seg-class}). There is no contradiction: a gain on the train class contributes one of 19 terms to aggregate mIoU even if it occurs in scenes outside the lowest fifth. Moreover, the images in that fifth can differ between models. The measured aggregate improvement therefore does not establish a benefit for difficult scenes or identify which search component produced the change.

\section{Discussion}

The two experiments agree on one practical point despite using different policy actions and rewards: small differences under shared weights were insufficient to select the eventual winner. In classification, the best independently trained finalist was tied with another candidate under the supernet. In segmentation, the selected cell ranked last on the shared-weight shortlist and did not lead after the first independent probe. These examples favor using shared weights to narrow a pool and reserving part of the budget to train finalists independently. They do not show how often this ordering would change in a new search.

With respect to model size, the classification and segmentation comparisons lead to different conclusions. The 86.94M-parameter classifier has fewer parameters than large NAONet, yet it is considerably larger than several methods with similar reported errors (Table~\ref{tab:comparison}). By contrast, the selected segmentation cell improves aggregate mIoU over reproduced Auto-DeepLab-S with 26,240 fewer parameters under matched training. In neither case does the parameter count alone establish inference speed, since operations with similar numbers of weights can have different computational costs.

The timing results in the paper describe search and training on the stated hardware. Without measurements of forward-pass computation, synchronized inference latency, or peak memory at a specified input size and batch size, they cannot establish deployment efficiency. Likewise, the classifier's SWA accuracy of 97.80\% describes one set of averaged weights; its position below the best individual checkpoint does not establish a general disadvantage of weight averaging.

Finally, independent training addresses the effect of sharing weights, but one seed does not quantify variability across initializations or searches. Repeating both stages across seeds would distinguish variation in the selected architecture from variation when training the same architecture again. For the classifier, selecting checkpoints on a separate validation split and evaluating the official test set only once would also separate checkpoint selection from final evaluation.

\section{Conclusion}

Both experiments use input-specific low-rank weight sharing and PPO, but the classification policy proposes moves in NAO's continuous representation, whereas the segmentation policy chooses cells directly and receives measured image-based rewards. In the classification experiment, the selected architecture reaches 2.00\% peak test error with 86.94M non-auxiliary parameters. In the matched segmentation experiment, changing one edge raises best single-scale validation mIoU from 73.942\% for reproduced Auto-DeepLab-S to 74.285\% under matched training, while the image-tail mean declines. Taken together, the results support independent evaluation of shortlisted cells and separate reporting of performance across classes and across difficult images; neither comparison by itself isolates the contribution of PPO or the adapters.

\end{document}